\documentclass[twoside,11pt]{article}

\usepackage{blindtext}

\usepackage{jmlr2e}

\usepackage{booktabs}
\usepackage{pifont}
\usepackage{subfig}
\usepackage{amsmath, bm}
\usepackage{multirow}
\usepackage{listings} 
\usepackage{xcolor}
\usepackage[figuresleft]{rotating}
\usepackage[T1]{fontenc}
\hypersetup{
    colorlinks=True,
    linkcolor=blue,
    filecolor=blue,      
    urlcolor=black,
    citecolor=blue,
}

\usepackage{lastpage}
\ShortHeadings{XuanCe: A Comprehensive and Unified Deep Reinforcement Learning Library}{Liu, Cai, Jiang, Cheng, Wang, Wang, Cao, Xu, Mu, and Sun}
\firstpageno{1}

\begin{document}

\title{XuanCe: A Comprehensive and Unified Deep Reinforcement Learning Library}

\author{\name Wenzhang Liu$^{1, 5, 6}$ \email wzliu@ahu.edu.cn \\
		\name Wenzhe Cai$^{2, 6}$ \email wz\_cai@seu.edu.cn \\
		\name Kun Jiang$^{2}$ \email kjiang@seu.edu.cn \\
		\name Guangran Cheng$^{2}$ \email chenggr@seu.edu.cn\\
		\name Yuanda Wang$^{2}$ \email wangyd@seu.edu.cn \\
		\name Jiawei Wang$^{3}$ \email wangjw@tongji.edu.cn \\
		\name Jingyu Cao$^{2}$ \email cjy0564@seu.edu.cn \\
		\name Lele Xu$^{2}$ \email xulele@seu.edu.cn \\
		\name Chaoxu Mu$^{4, 5, 6}$ \email cxmu@tju.edu.cn\\
		\name Changyin Sun$^{1, 5}$ (corresponding author) \email cysun@seu.edu.cn\\
		\addr $^{1}$School of Artificial Intelligence, Anhui University, Hefei, 230601, China \\
		\addr $^{2}$School of Automation, Southeast University, Nanjing, 210096, China \\
		\addr $^{3}$College of Electronics and Information Engineering, Tongji University, Shanghai, 201804, China\\
		\addr $^{4}$School of Electrical and Automation Engineering, Tianjin University, Tianjin, 300072, China\\
		\addr $^{5}$Engineering Research Center of Autonomous Unmanned System Technology, Ministry of Education, Hefei, 230601, China\\
		\addr $^{6}$Peng Cheng Laboratory, Shenzhen, 518000, China}
\editor{My editor}

\maketitle

\begin{abstract}%   <- trailing '%' for backward compatibility of .sty file
In this paper, we present XuanCe, a comprehensive and unified deep reinforcement learning (DRL) library designed to be compatible with PyTorch, TensorFlow, and MindSpore. XuanCe offers a wide range of functionalities, including over 40 classical DRL and multi-agent DRL algorithms, with the flexibility to easily incorporate new algorithms and environments. It is a versatile DRL library that supports CPU, GPU, and Ascend, and can be executed on various operating systems such as Ubuntu, Windows, MacOS, and EulerOS. Extensive benchmarks conducted on popular environments including MuJoCo, Atari, and StarCraftII multi-agent challenge demonstrate the library's impressive performance. XuanCe is open-source and can be accessed at \href{https://github.com/agi-brain/xuance}{https://github.com/agi-brain/xuance.git}.
\end{abstract}


\section{Introduction}

In recent years, deep reinforcement learning (DRL) has achieved many advances in various applications, such as video games \citep{mnih2015human}, board games \citep{silver2018general}, and intelligent chat \citep{radford2019language}, etc. 
The success of DRL can largely be attributed to the development of deep learning (DL) \citep{lecun2015deep}. Nowadays, several toolboxes such as PyTorch~\citep{paszke2019pytorch}, TensorFlow~\citep{abadi2016tensorflow}, and MindSpore~\citep{huawei2022huawei}, etc., have been open-sourced for people to implement DL applications. With the growing interest in DRL algorithms for intelligent decision making, a DRL library that supports these DL toolboxes is urgently needed. 

Unlike supervised or unsupervised learning, the training of DRL agents involves collecting raw data through interactions with the environments, and training the policy by maximizing accumulated rewards. The design of DRL algorithms typically depends on the specific environments and tasks, which brings challenges in unifying various algorithms within one architecture. Additionally, ensuring the reusability and extensibility of the APIs for different types of DRL agents proves to be difficult. Moreover, compatibility issues arise when using different DL toolboxes. Hence, building a unified and comprehensive library that encompasses various types of DRL algorithms while supporting multiple DL toolboxes poses a significant challenge.

In this paper, we introduce XuanCe, a DRL library that includes more than 40 DRL and multi-agent reinforcement learning (MARL) algorithms. This library is compatible with PyTorch, TensorFlow, and MindSpore. It supports CPU, GPU, Ascend, and can be run on the Ubuntu, Windows, MacOS, and EulerOS. With the modularized components and unified APIs, it is easier for users to develop new algorithms and add new environments. The benchmark results have verified the performance of the XuanCe library.

\section{Related Work}

Recently, numerous open-sourced repositories that offer a diverse array of DRL algorithms have been proposed. 
RLlib~\citep{liang2018rllib}, along with its successors FinRL~\citep{liu2021finrl} and OR-Gym~\citep{hubbs2020or}, introduces a parallel training framework catering to both DRL and MARL. However, the highly encapsulated modules in these frameworks could potentially constrain their flexibility and extensibility.
Dopamine~\citep{castro18dopamine} focuses specifically on algorithms based on deep Q-networks (DQN), whereas SpinningUp~\citep{SpinningUp2018} specializes in algorithms based on policy gradients (PG). These libraries prioritize specific algorithm families, which may lead to a narrower scope of applicability.
To address compatibility and readability concerns, some libraries are designed with highly modular structures, such as Tonic~\citep{pardo2020tonic} and Tianshou~\citep{weng2022tianshou}. 
RLzoo~\citep{ding2020rlzoo} and ChainerRL~\citep{fujita2021chainerrl} are designed specifically for the TensorLayer and Chainer frameworks, respectively, thereby constraining their versatility.
d3rlpy~\citep{seno2022d3rlpy} primarily focuses on the offline DRL algorithms, which may restrict its application in online and interactive learning scenarios.
MushroomRL~\citep{JMLR:v22:18-056} stands out by providing a rich set of components that encompass the requirements of typical DRL algorithms. It simplifies implementation and testing through its benchmarking suite, offering a practical and comprehensive solution.
skrl~\citep{serrano2023skrl} is also designed with high modularity to ensure the readability, simplicity, and transparency of algorithms.
To address the compatibility issues and improve the scalability for MARL library, MARLlib~\citep{hu2023marllib} provides standardized wrappers, agent-level implementations, and flexible policy mapping strategies.

The proposed XuanCe incorporates the strengths of the aforementioned libraries, and offers a diverse range of algorithms including both single-agent DRL and MARL for different tasks. By employing a unified framework and modular design, XuanCe allows users to quickly comprehend and train their specific tasks.

\section{Design of XuanCe}

\begin{figure}
	\centering
	\includegraphics[width=1.0\columnwidth]{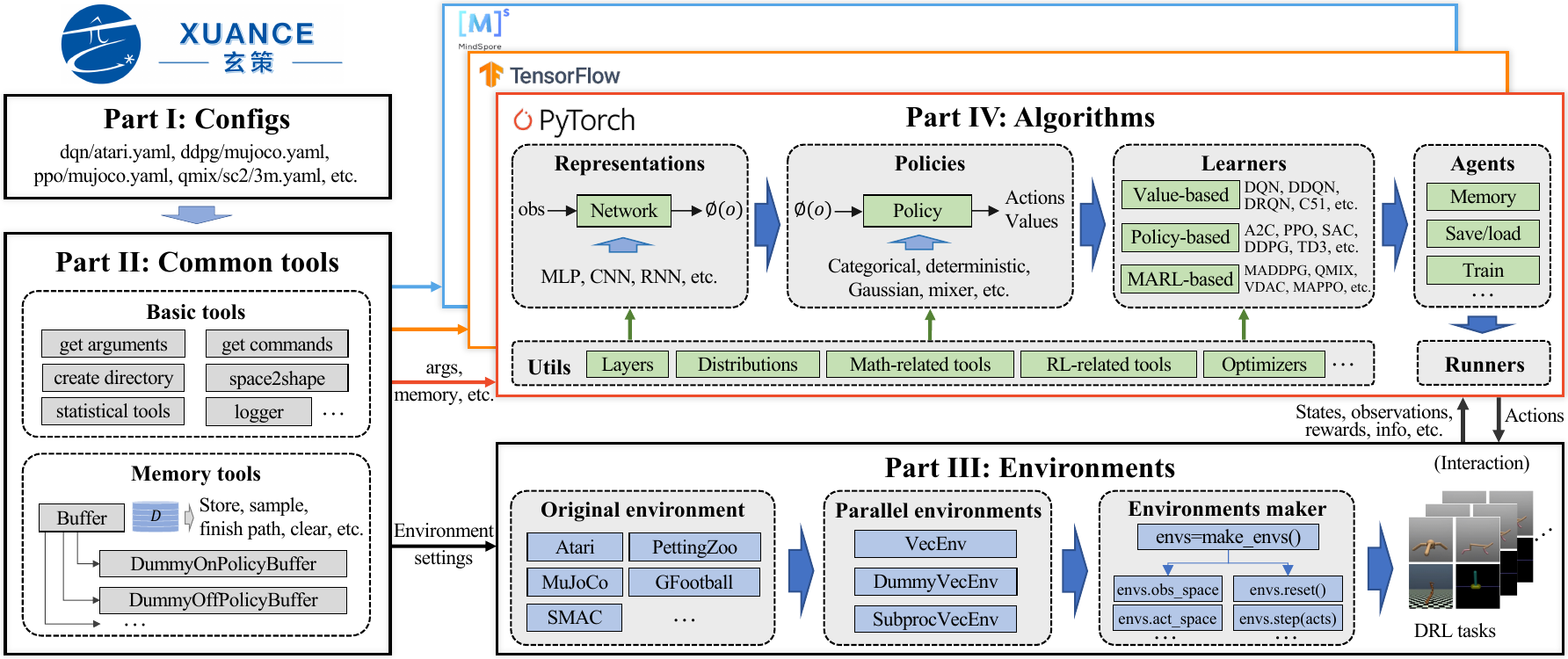}
	\caption{The framework of XuanCe DRL library.}
	\label{fig:xuance-framework}
\end{figure}

The overall framework of XuanCe is shown in Figure \ref{fig:xuance-framework}. XuanCe contains four main parts: configs, common tools, environments, and algorithms.

\subsection{Configs}\label{sec:design:configs}

To simplify the configuration of hyper-parameters, environments, models, etc., we utilize YAML files to assign the arguments for each implementation. In the config file, we can select the name of environment, representation, policy, etc., and set the values such as buffer size, learning rate, discount factor, etc. This flexibility enables users to define any necessary attributes for the implementation.

\subsection{Common Tools}

Within this part, various reusable tools are developed, which are independent of the choice of DL toolbox. These tools encompass both basic tools and memory tools. The basic tools serve to prepare the DRL model before training, such as loading hyper-parameters from YAML files, getting terminal commands, and creating model directories, etc. The memory tools are employed to construct a replay buffer that collects experiences data during the interaction between the agent(s) and environments.

\subsection{Environments}

XuanCe supports numerous simulated tasks in the part of environments. To improve the sample efficiency, we create parallel environments based on the original environment, and standardize the APIs for interaction. For MARL with cooperative and competitive tasks, we group the agents in the environment according to their roles. Users can also extend their own environment by wrapping the original environment with same APIs of XuanCe. After that, we can easily make the environments for experience collection and model training.

\subsection{Algorithms}

The key part of XuanCe is algorithms. For each DL toolbox, this part provides five unified modules: utils, representations, policies, learners, agents, and runners. 

\textbf{Utils}. This module contains utilities used to construct fundamental neural layers, stochastic distributions, mathematical tools, DRL tools, and optimizers, etc. These tools are customized for the specific DL toolbox and are reusable across other modules, including representations, policies, and learners.

\textbf{Representations}. Learning a good representation is important for the decision making of a DRL agent. In this module, we can build a neural network for agents to extract features from observation data. The choices in the representations module include multi-perceptron layer (MLP), convolutional neural networks (CNN), recurrent neural networks (RNN), etc. The users can also customize the networks for their specific scenarios.

\textbf{Policies}. To generate actions and values for agents to interact with the environment, this module provides various types of policies, such as categorical policies, deterministic policies, and Gaussian policies, etc. The categorical and Gaussian policies are designed for discrete and continuous actions, respectively. They generate actions via sampling from stochastic distributions. Different from that, deterministic policies directly generate executable actions for agents. All policies take the outputs of the representations as their inputs, thus concluding the feedforward processes. Users can select or customize the Q-networks, actor networks, critic networks, and their target duplications, etc., in this module.

\textbf{Learners}. For each DRL algorithm, we design a learner to update the parameters of the representation and the policy. This module contains three kinds of learners: value-based, policy-based, and MARL-based learners. The update method in each learner is mainly responsible for calculating the loss, performing gradient descent on trainable parameters, and updating target networks. Hence, it is the key module of a DRL algorithm.

\textbf{Agents}. In this module, we create a memory for experiences collecting, a learner for parameters updating, and a method for actions generating. Then, we define methods to load the pre-trained models, train the models, and save the models, etc. The agents module is also divided into three types: value-based, policy-based, and MARL-based agents, each corresponding one-to-one with the learners module. 

\textbf{Runners}. The runners module provides high-level APIs for DRL implementation. Within this module, we create an agent to interact with the environments, collect and log the training data for subsequent analysis. Users also have the flexibility to customize the runner file to execute their implementations.

The testing results for various benchmark tasks using the algorithms in XuanCe can be found in the appendix \ref{sec:appendix:C}, providing evidence of the library's quality.

\section{Conclusion}

In this paper, we introduce an open-sourced DRL library called XuanCe, which is compatible with PyTorch, TensorFlow, and MindSpore. XuanCe is highly modularized and easy to install and use for researchers. It currently support more than 40 algorithms, including single agent DRL and MARL, and is easy for users to extend new algorithms. The experiment results on some popular environments have verified the quality of the library. Full documentation of XuanCe is available at \href{http://xuance.readthedocs.io/}{http://xuance.readthedocs.io/}.

\newpage

\section*{Acknowledgement}
We thank all the open source contributors for their help on the development. This work is supported by the National Natural Science Foundation of China (Nos. 62236002, 61921004, and 62103104). We also acknowledge the High-performance Computing Platform of Anhui University for providing computing resources.

% Manual newpage inserted to improve layout of sample file - not
% needed in general before appendices/bibliography.

%\vskip 0.2in
%\bibliography{reference.bib}

\appendix

\section{Usage}

\subsection{Installation}

This library can be installed from Python Package Index (PyPI), a widely-used repository of software for the Python programming language. Alternatively, it can also be installed offline after being download from GitHub. Here are two examples of how to install XuanCe.

\begin{lstlisting}
	$ # prepare conda environment (Anaconda3 needs to be installed)
	$ conda create -n xuance_env python=3.7
	$ conda activate xuance_env
	
	$ # Method 1: Install XuanCe from PyPI.
	$ pip install xuance
	
	$ # Method 2: Install XuanCe offline.
	$ git clone https://github.com/agi-brain/xuance.git
	$ cd xuance
	$ pip install -e .
\end{lstlisting}

It should be noted that some other libraries should be installed for users with different requirements. For example, PyTorch, Atari environment, MuJoCo simulator, etc. More information about the installation of the library can be found in its documentation\footnote{\href{https://xuance.readthedocs.io/en/latest/documents/usage/installation.html}{https://xuance.readthedocs.io/en/latest/documents/usage/installation.html}}.

\subsection{Basic Usage}\label{sec:Usage:basic-usage}

After installing the library, we can run the a DRL algorithm with just three lines of code by specifying the names of algorithm and environment. Here is a concise usage demo for DQN algorithm solving "CartPole-v1", a typical task of the classic control environment in OpenAI Gym~\citep{brockman2016openai}.
\begin{lstlisting}
	1. import xuance
	2. runner = xuance.get_runner(method="dqn",
			              env="classic_control",
			              env_id="CartPole-v1",
			              is_test=False)
	3. runner.run()  # start running the algorithm
\end{lstlisting}

The parameters of the trained model are stored in files with ".pth" extension, which is located in the "./models/dqn/torch/CartPole-v1/" directory. The training results are stored in files that are stored in the "./logs/dqn/torch/CartPole-v1/" directory, and can be visualized by the tensorboard or wandb tools. More examples could be found in the Github page: \href{https://github.com/agi-brain/xuance/tree/master/examples}{https://github.com/agi-brain/xuance/tree/master/examples}.

\section{Advantages of XuanCe}

\subsection{Supported Algorithms}

XuanCe currently supports various types of DRL algorithms, such as value-based algorithms for discrete action spaces, policy-based algorithms for both discrete and continuous action spaces, and MARL-based algorithms for cooperative and competitive multi-agent tasks. All of the algorithms in XuanCe are implemented under a unified framework (Figure \ref{fig:xuance-framework}), and they can be run with PyTorch, TensorFlow2, and MindSpore. Users can apply the algorithms of XuanCe to solve different tasks by selecting the suitable representation module, policy module, learner module, and other parameters. Table \ref{table:algorithm-list} lists the currently implemented algorithms. 

\begin{table}[h] \small
\caption{The algorithms that are currently implemented in XuanCe}
\centering
\begin{tabular}{c|l}
\toprule
\textbf{Type}	  & \textbf{Implemented algorithms} \\
\midrule
				  & Deep Q-Network (DQN), Double DQN, Duel DQN, \\
Value-based		  & C51DQN, Noisy DQN, Deep Recurrent Q-Network (DRQN), \\
(single-agent)	  & Prioritized Experience Replay DQN (PER-DQN),  \\
				  & Quantile Regression DQN (QRDQN), Parameterized DQN (PDQN), \\
				  & Split-PDQN (SPDQN), Multi-pass DQN (MPDQN). \\
\midrule
				  & Policy Gradient (PG), Phasic PG (PPG), Advantage Actor-Critic (A2C), \\
Policy-based		  & Deep Deterministic PG (DDPG), Twin-Delayed DDPG (TD3), \\
(single-agent) 	  & Soft Actor-Critic (SAC), SAC with discrete actions spaces (SAC-Dis),\\
				  & Proximal Policy Optimization (PPO), PPO with KL divergence (PPO-KL). \\
\midrule
		          & Independent Q-Learning (IQL), Value-Decomposition Network (VDN, VDN-S), \\
				  & QMIX, Weighted QMIX (OWQMIX, CWQMIX), \\
				  & QTRAN (QTRAN-base, QTRAN-alt), \\
				  & Deep Coordination Graph (DCG, DCG-S), \\
MARL-based		  & Independent DDPG (IDDPG), Multi-Agent DDPG (MADDPG), \\
				  & Independent SAC (ISAC), Multi-Agent SAC (MASAC), \\
				  & Multi-Agent TD3 (MATD3), \\
				  & Mean-Field Q-Learning (MFQ), Mean-Field Actor-Critic (MFAC), \\
				  & Counterfactual Multi-Agent Policy Gradient (COMA), \\
				  & Value-Decomposition Actor-Critic (VDAC-sum, VDAC-mix), \\
				  & Independent PPO (IPPO), Multi-Agent PPO (MAPPO). \\
\bottomrule
\end{tabular}
\label{table:algorithm-list}
\end{table}

\subsection{Supported Environments}

\begin{figure}
	\centering
	\subfloat[]{
		\includegraphics[width=0.22\columnwidth]{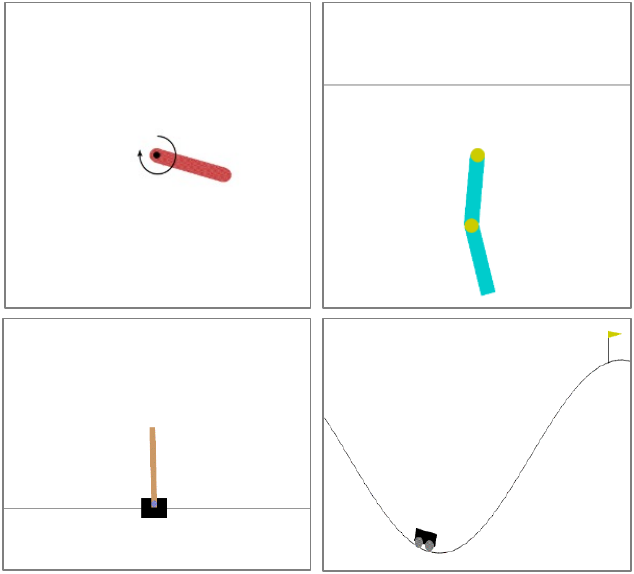}
		\label{fig:drl-framework}
	}
	\subfloat[]{
		\includegraphics[width=0.20\columnwidth]{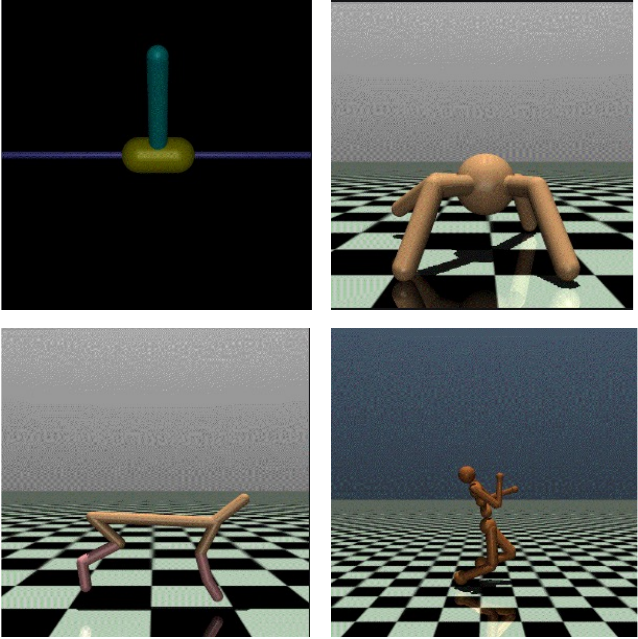}
		\label{fig:drl-framework}
	}
	\subfloat[]{
		\includegraphics[width=0.20\columnwidth, height=0.20\columnwidth]{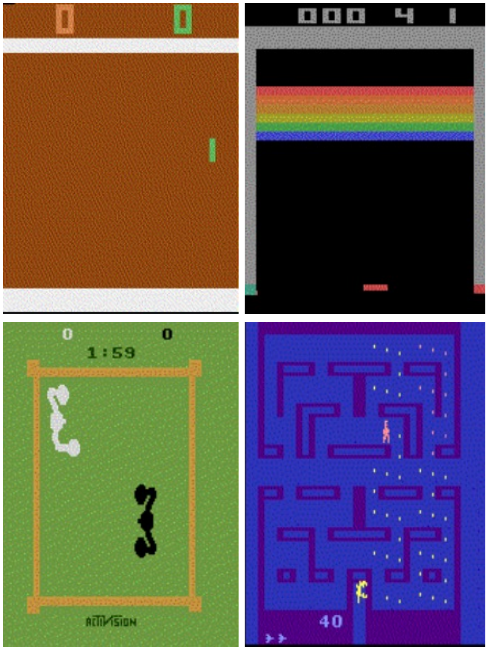}
		\label{fig:drl-framework}
	}
	\subfloat[]{
		\includegraphics[width=0.32\columnwidth]{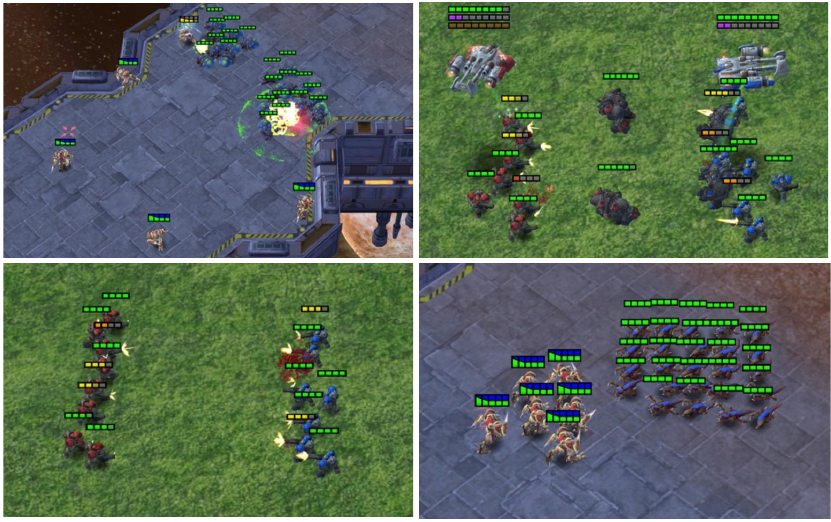}
		\label{fig:drl-framework}
	}
	\caption{Part of the supported single-agent and multi-agent environments: (a) Classic Control, (b) MuJoCo, (c) Atari, (d) SMAC.}
	\label{fig:environments}
\end{figure}

This library supports widely-used environments in DRL research community, including Classic Control, Box2D, MuJoCo, Atari, etc. These environments are part of the gym library~\citep{brockman2016openai}, which is easy to install. Additionally, the library also supports multi-agent environments with cooperative and/or competitive tasks, such as StarCraftII Multi-Agent Challlenge (SMAC)~\citep{vinyals2017starcraft}, Google Research Football (GRF)~\citep{kurach2020google}, MAgent2~\citep{zheng2018magent}, and so on. Part of the supported environments is illustrated in Figure \ref{fig:environments}. 

\subsection{Extensibility for New Algorithms}

It is easy for users to develop new algorithms in XuanCe library. As shown in Figure~\ref{fig:xuance-framework}, the library contains most of the fundamental components for a DRL algorithm. Hence, the users should add the following key components to develop a new DRL algorithm: 
\begin{itemize} 
	\item Configuration file. To begin the design of a new algorithm in XuanCe, we first need to create a YAML file with required arguments. As introduced in section \ref{sec:design:configs}, the configuration file should be used to set values for hyper-parameters, select environment, representation, policy, and other modules.
	\item Learner file. Subsequently, a learner file for the new algorithm should be created. This file contains the definition of the "update" method, which is the core component responsible for calculating losses and updating the model's parameters. 
	\item Agent file. Finally, we need to create a new agent file, in which a memory module, a learner module, and a method for generating actions are created. In this file, we also need to define the methods for model training, saving, loading, and testing.
\end{itemize}
After adding these files, we need update the "\_\_init\_\_.py" files of the learners and agents modules, such that the new items can be successfully imported. 
It should be noted that, these are the basic steps for implementing a new algorithm. If any module of the new algorithm cannot be matched in XuanCe, such as the representation, policy, or memory, it should also be added to the corresponding file. 
After that, users can then run the new algorithm by following the steps outlined in section \ref{sec:Usage:basic-usage}. 

\subsection{Extensibility for New Environments}

Extending a new environment for customized implementations is a straightforward process. To ensure compatibility with XuanCe, it is necessary to create a new folder (e.g., "xuance/environments/new\_env/") that includes two new files: one with the implementation of the new environment (e.g., "new\_env.py"), and another responsible for vectorizing the environment for parallel execution (e.g., "new\_vec\_env.py"). It is important to note that the APIs of both original environment and parallel environments should be ensured to be consistency with XuanCe. In the library's environment module, we have included an example of a new environment implementation located in the "/xuance/environment/new\_env/" folder. 
After creating the original environment and parallel environments for new implementation, we can import and make the environments in the runner module and enable the agent-environment interactions.

\subsection{XuanCe vs Other Libraries}

Although there are several DRL libraries being open-source in recent years, they often fall short in terms of simplicity, compatibility and the variety of algorithms, making them unable to fully meet the diverse needs of different users. For example, RLlib~\citep{liang2018rllib} highly encapsulates the APIs, increasing the complexity of the framework and reducing the flexibility for extensions. Consequently, it is challenging for DRL beginners to quickly get started. Tianshou~\citep{weng2022tianshou} and MARLlib~\citep{hu2023marllib} improve the simplicity and readability via highly modularized design. However, they support either single-agent DRL algorithms or MARL algorithms. Another modularized library, skrl~\citep{serrano2023skrl}, supports both single-agent DRL and MARL. However, it has a limited number of algorithms, with currently only IPPO and MAPPO available for MARL.

\begin{table}[ht]\small
\caption{Comparison of XuanCe and other popular libraries}
\centering
\begin{tabular}{lccccc}
\toprule
\textbf{DRL library} & \textbf{PyTorch} & \textbf{TensorFlow} & \textbf{MindSpore} & {\textbf{MARL}} & {\textbf{\#} \textbf{Algo.}} \\
\midrule
RLlib%~\citep{liang2018rllib}
& \checkmark & \checkmark & \ding{55} & \checkmark	& 34 \\
SpinningUp%~\citep{SpinningUp2018}
& \checkmark & \checkmark & \ding{55} & \ding{55}	& 6 \\
Dopamine%~\citep{castro18dopamine}
& \ding{55}  & \checkmark& \ding{55} & \ding{55}	& 4 \\
Tonic%~\citep{pardo2020tonic}
& \checkmark & \checkmark & \ding{55} & \ding{55}	& 10 \\
RLzoo%~\citep{ding2020rlzoo}
& \checkmark & \checkmark & \ding{55} & \ding{55}	& 14 \\
MushroomRL%~\citep{JMLR:v22:18-056}
& \checkmark & \ding{55} & \ding{55} & \ding{55}	& 33 \\
d3rlpy%~\citep{seno2022d3rlpy}
& \checkmark & \ding{55} & \ding{55} & \ding{55}	& 27 \\
ChainerRL%~\citep{fujita2021chainerrl}
& \ding{55}  & \ding{55}  & \ding{55} & \ding{55}	& 24 \\
Tianshou%~\citep{weng2022tianshou}
& \checkmark & \ding{55} & \ding{55} & \ding{55}	& 32 \\
FinRL%~\citep{liu2021finrl}
& \checkmark & \checkmark & \ding{55} & \ding{55}	& 5 \\
skrl%~\citep{serrano2023skrl}
& \checkmark	& \ding{55}	& \ding{55} & \checkmark	& 15 \\
MARLlib%~\citep{hu2023marllib}
& \checkmark	 & \ding{55} & \ding{55} & \checkmark	& 18 \\
\textbf{XuanCe} &
\checkmark & \checkmark & \checkmark & \checkmark	&	\textbf{42}	\\
\bottomrule
\end{tabular}
\label{table:comparison:libraries}
\end{table}

XuanCe distinguishes itself by providing a comprehensive set of components that cater to the requirements of various DRL algorithms for both single-agent and multi-agent systems, and simplifying the implementations through a highly modularized design. It provides a unified framework that supports three kinds of DL toolboxes: PyTorch, TensorFlow, and MindSpore. MindSpore is a DL toolbox for "device-edge-cloud" scenarios, with the goal of bridging the gap between algorithm research and production deployment. This aligns with the design motivation of XuanCe. Hence, we choose to expand the compatibility of XuanCe beyond the widely-used PyTorch and TensorFlow to include the MindSpore. 
Table~\ref{table:comparison:libraries} lists the comparison of XuanCe and other popular libraries.

\section{Benchmark Results}\label{sec:appendix:C}

In this section, we conducted an evaluation of 14 commonly used algorithms in XuanCe on 78 different tasks from three widely-used environments: MuJoCo, Atari, and SMAC. The evaluation was conducted on a desktop computer with 64GB RAM, one Intel Core i9-12900K CPU, and one Nvidia GeForce RTX 3090Ti GPU.

\subsection{MuJoCo}

The reproduced MuJoCo environment is a component of the OpenAI Gym library, providing 10 distinct continuous control tasks, including robotic control, manipulation, and locomotion, etc. It is a physics engine standing for multi-joint dynamics with contract. To evaluate the performance of XuanCe on this environment, we have selected both on-policy and off-policy algorithms capable of handling continuous action spaces. The on-policy algorithms include PG, A2C, and PPO, while the off-policy algorithms include DDPG, TD3, and SAC.

\begin{table}[htbp]\tiny
\caption{Benchmark results of MuJoCo for on-policy algorithms in XuanCe}
\centering
\begin{tabular}{l|cc|cc|cc}
\toprule
		& \multicolumn{2}{c|}{\textbf{PG}} & \multicolumn{2}{c|}{\textbf{A2C}} & \multicolumn{2}{c}{\textbf{PPO}}  \\
\midrule
\textbf{Task} & \textbf{XuanCe} & \textbf{Pub.} & \textbf{XuanCe} & \textbf{Pub.} & \textbf{XuanCe} & \textbf{Pub.} \\
\midrule
Ant 	 	& 775.17$\pm$29.55 & - & 1931.02$\pm$290.26 & - 	  & 3862.00$\pm$572.12 & - \\
HalfCheetah & 1008.61$\pm$143.20 & - & \textbf{1230.59$\pm$604.03} & \textasciitilde 1000   & \textbf{4778.96$\pm$511.99} & \textasciitilde 1800 \\
Hopper		& 423.23$\pm$189.95   & - & 643.48$\pm$285.32 & \textbf{\textasciitilde 900} 	& \textbf{3185.79$\pm$173.05} & \textasciitilde 2330 \\
HumanoidStandup	& 92668.28$\pm$20337.80 & - & 114717.93$\pm$20751.35 & - & 96047.95$\pm$13962.73 & - \\
Humanoid	& 351.73$\pm$45.81 & - & 731.99$\pm$196.24  & - & 490.84$\pm$30.80  & - \\
InvDoublePendulum	& 693.41$\pm$114.56 & - & \textbf{9342.31$\pm$4.64} & \textasciitilde 8100   & \textbf{9350.91$\pm$3.36} & \textasciitilde 8000 \\
InvPendulum	& 1000.0$\pm$0.0 & - & \textbf{1000.0$\pm$0.0} & \textbf{\textasciitilde 1000.0} & \textbf{1000.0$\pm$0.0} & \textbf{\textasciitilde 1000.0} \\
Reacher		& -11.70$\pm$0.76 & - & \textbf{-12.92$\pm$3.06} & \textasciitilde -24 & \textbf{-5.54$\pm$0.44} & \textasciitilde -7 \\
Swimmer		& 39.14$\pm$1.77  & - & \textbf{38.58$\pm$4.94}   & \textasciitilde 31 	  & 102.20$\pm$14.95  & \textbf{\textasciitilde 108} \\
Walker2d	& 699.13$\pm$121.71 & - & 773.88$\pm$241.83 & \textbf{\textasciitilde 850} & \textbf{3818.25$\pm$1270.77} & \textasciitilde 3460 \\
\midrule
\# Higher & - & - & 4 & 2 & 5 & 1 \\
\bottomrule
\end{tabular}
\label{table:benchmark:mujoco:on-policy}
\end{table}

\begin{table}[htbp]\tiny
\caption{Benchmark results of MuJoCo for off-policy algorithms in XuanCe}
\centering
\begin{tabular}{l|cc|cc|cc}
\toprule
		& \multicolumn{2}{c|}{\textbf{DDPG}} & \multicolumn{2}{c|}{\textbf{TD3}}& \multicolumn{2}{c}{\textbf{SAC}} \\
\midrule
\textbf{Task}   & \textbf{XuanCe} & \textbf{Pub.} & \textbf{XuanCe} & \textbf{Pub.} & \textbf{XuanCe} & \textbf{Pub.} \\
\midrule
Ant 	 	& \textbf{1411.91$\pm$309.85} & 1005.3 & \textbf{4866.89$\pm$270.32}  & 4372.44$\pm$1000.33 & \textbf{2303.14$\pm$258.57} & 655.35 \\
HalfCheetah & \textbf{10756.74$\pm$1075.15}  & 3305.6 & \textbf{10681.25$\pm$640.84} & 9636.75$\pm$859.07  & \textbf{7307.77$\pm$1273.87} & 2347.19 \\
Hopper		& \textbf{3689.74$\pm$14.12} & 2020.5 & \textbf{3782.21$\pm$88.22} & 3564.07$\pm$114.74  & 1810.42$\pm$1446.18 & \textbf{2996.66} \\
HumanoidStandup	& 83686.07$\pm$24915.99 & - & 81426.84$\pm$25097.73 & - & 144783.30$\pm$9709.48 & - \\
Humanoid	& 153.46$\pm$121.07	& - & 66.46$\pm$10.17 & - & 1566.68$\pm$445.61 & - \\
InvDoublePendulum	& 9343.02$\pm$22.58 & \textbf{9355.5} & \textbf{9347.34$\pm$10.98} & 9337.47$\pm$14.96 & \textbf{9359.77$\pm$0.22} & 8487.15 \\
InvPendulum	& \textbf{1000.0$\pm$0.0} & \textbf{1000.0} & \textbf{1000.0$\pm$0.0} & \textbf{1000.0$\pm$0.0} & \textbf{1000.0$\pm$0.0} & \textbf{1000.0} \\
Reacher		& \textbf{-4.88$\pm$0.34}  & -6.5   & -4.84$\pm$0.36 & \textbf{-3.60$\pm$0.56} 	   & -5.65$\pm$0.39 & \textbf{-4.44} \\
Swimmer		& 143.16$\pm$49.70   & - 	   & 72.00$\pm$34.78 	 & - 	   & 74.14$\pm$21.89   & - \\
Walker2d	& \textbf{4827.07$\pm$507.23} & 1843.6 & \textbf{5297.93$\pm$544.93} & 4682.82$\pm$539.64  & \textbf{3617.24$\pm$306.28} & 1283.67 \\
\midrule
\# Higher & 5 & 1 & 5 & 1 & 4 & 2 \\
\bottomrule
\end{tabular}
\label{table:benchmark:mujoco:off-policy}
\end{table}

\begin{figure}[h]
	\centering
	\subfloat[Ant]{
		\includegraphics[height=0.20\textwidth]{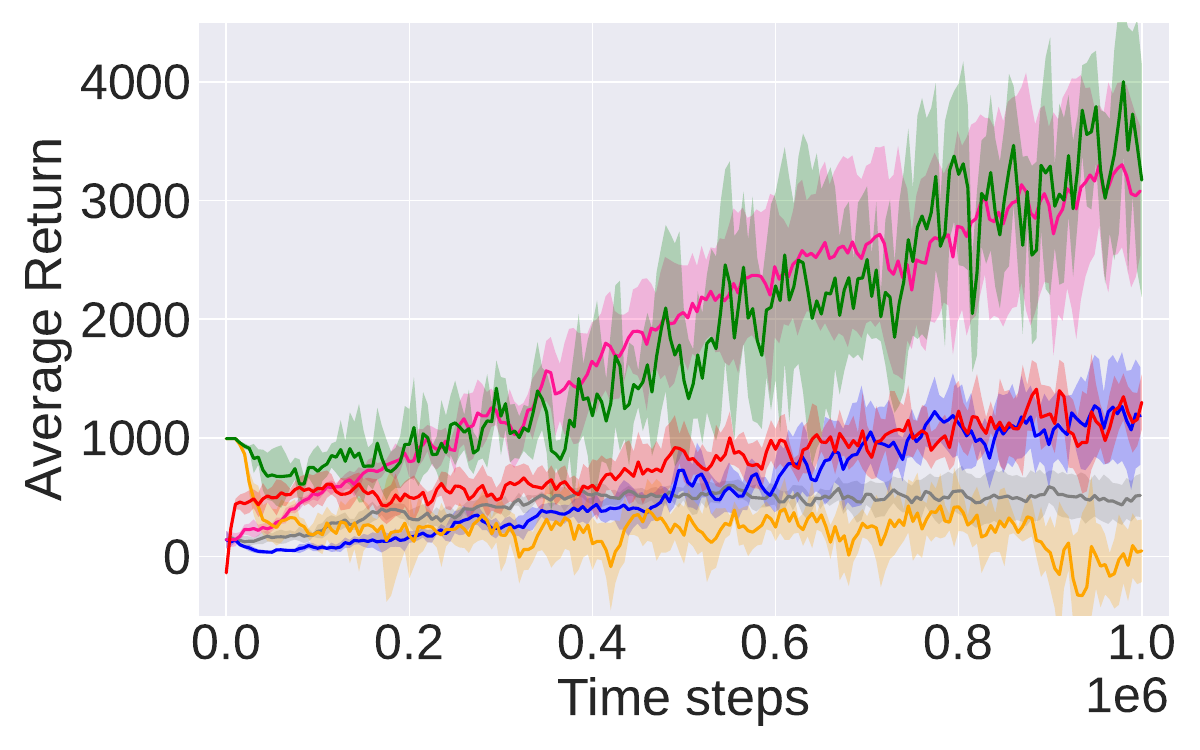}}
	\subfloat[HalfCheetah]{
		\includegraphics[height=0.20\textwidth]{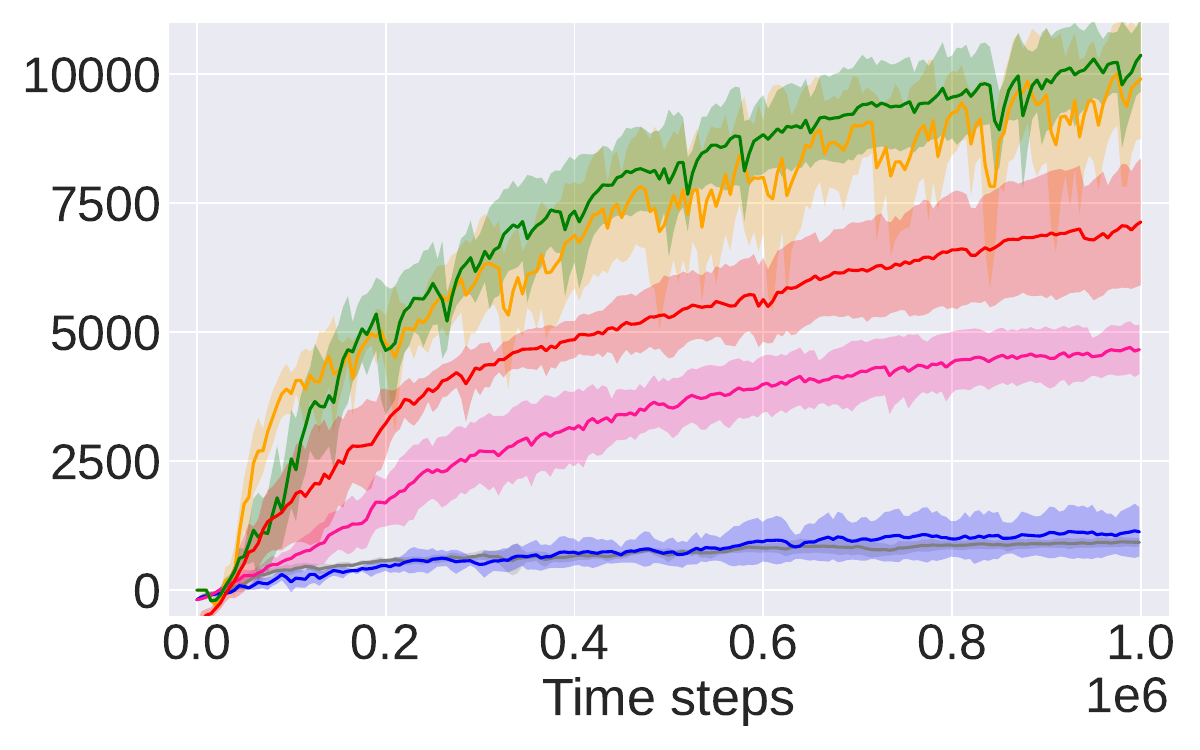}}
	\subfloat[Hopper]{
		\includegraphics[height=0.20\textwidth]{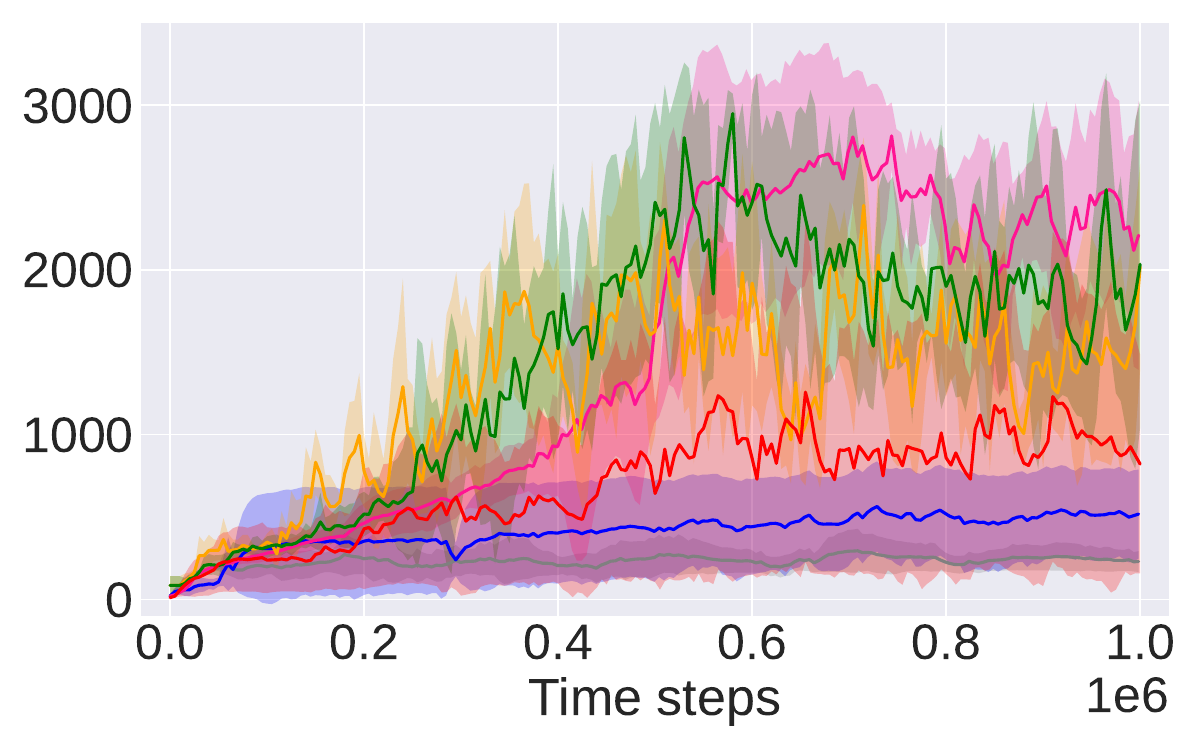}}
	\vfill
	\subfloat[HumanoidStandup]{
		\includegraphics[height=0.20\textwidth]{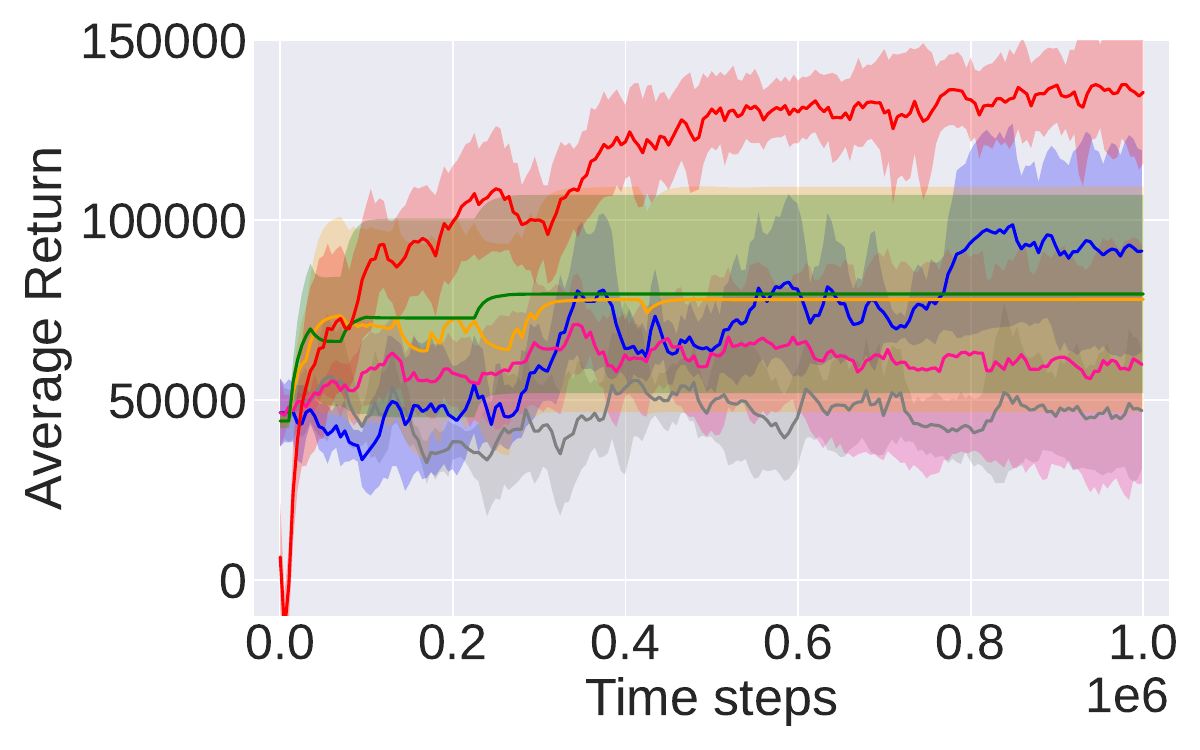}}
	\subfloat[Humanoid]{
		\includegraphics[height=0.20\textwidth]{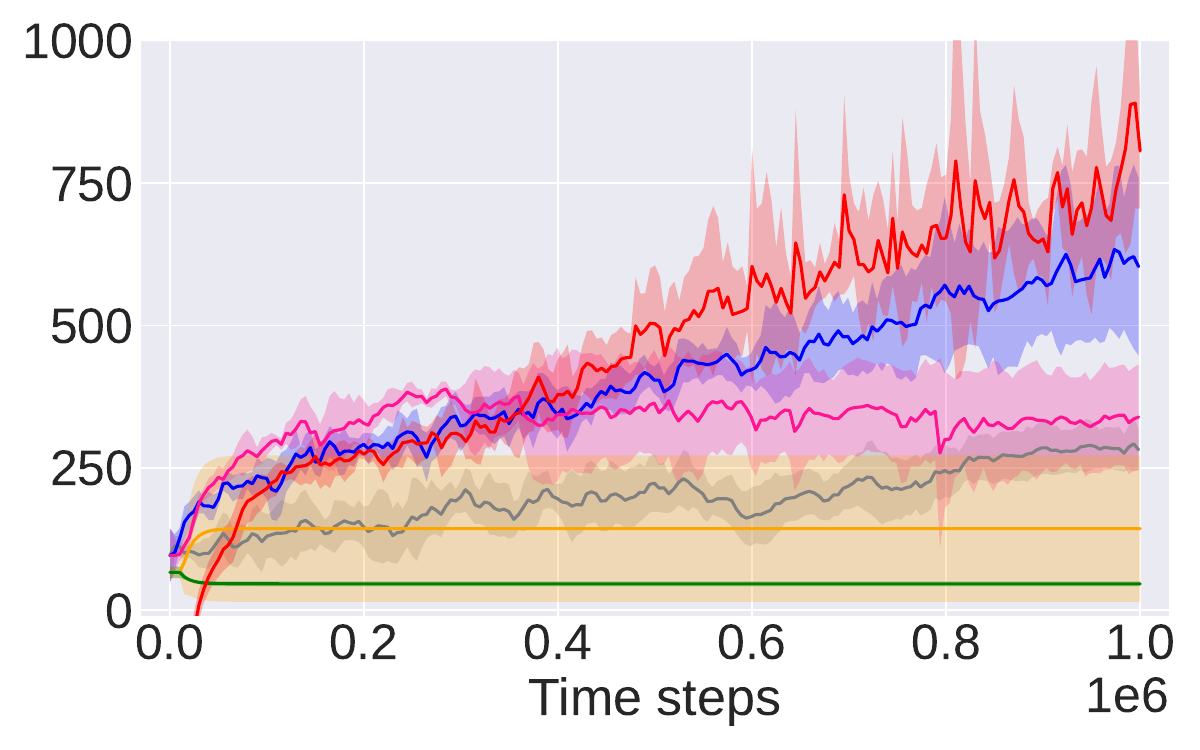}}
	\subfloat[InvDoublePendulum]{
		\includegraphics[height=0.20\textwidth]{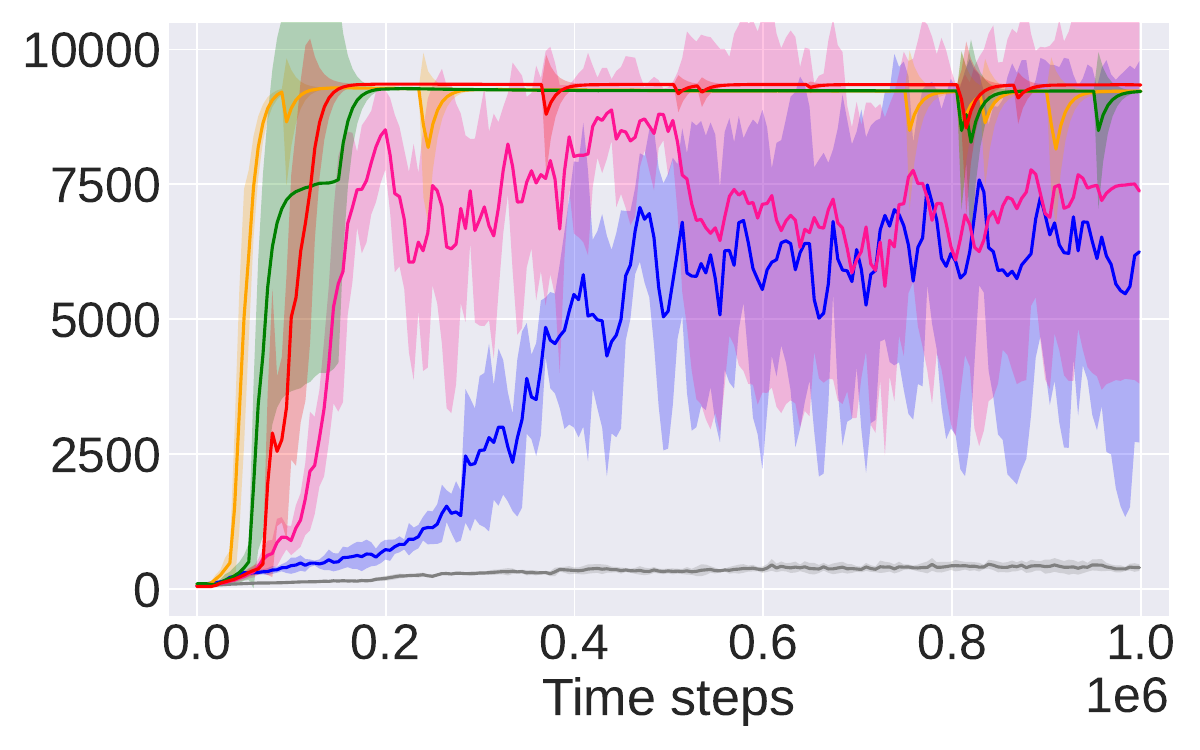}}
	\vfill
	\subfloat[InvPendulum]{
		\includegraphics[height=0.20\textwidth]{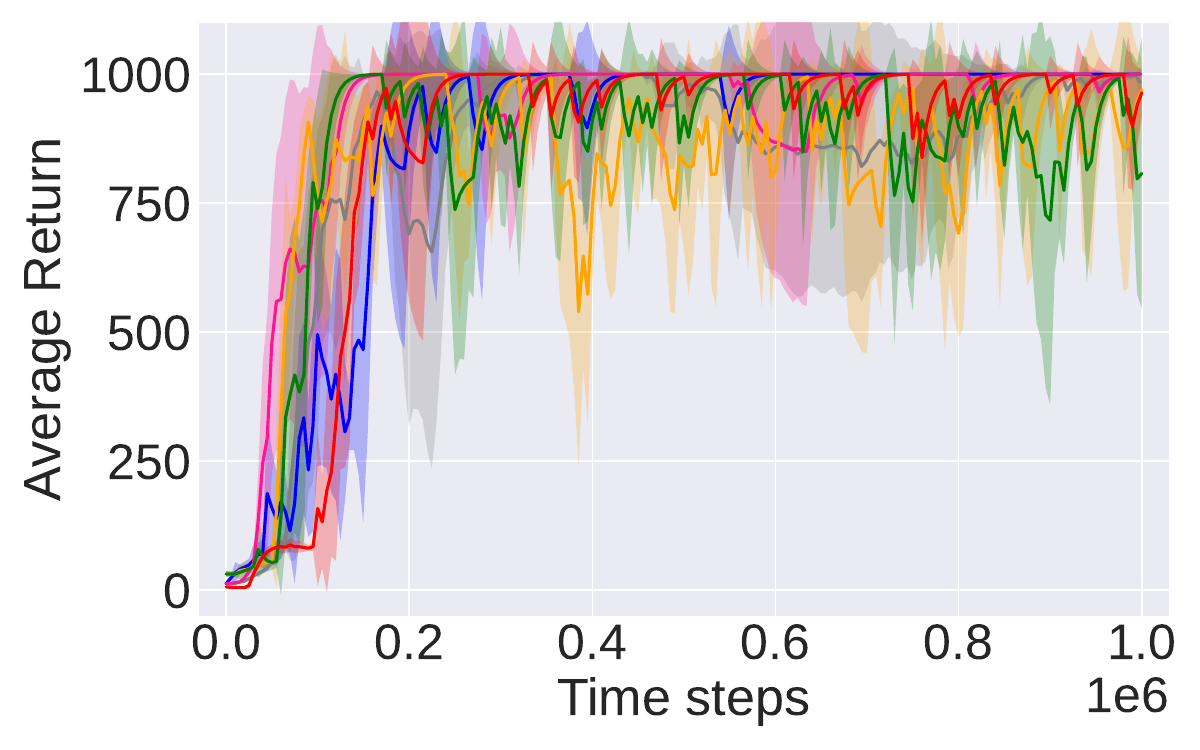}}
	\subfloat[Reacher]{
		\includegraphics[height=0.20\textwidth]{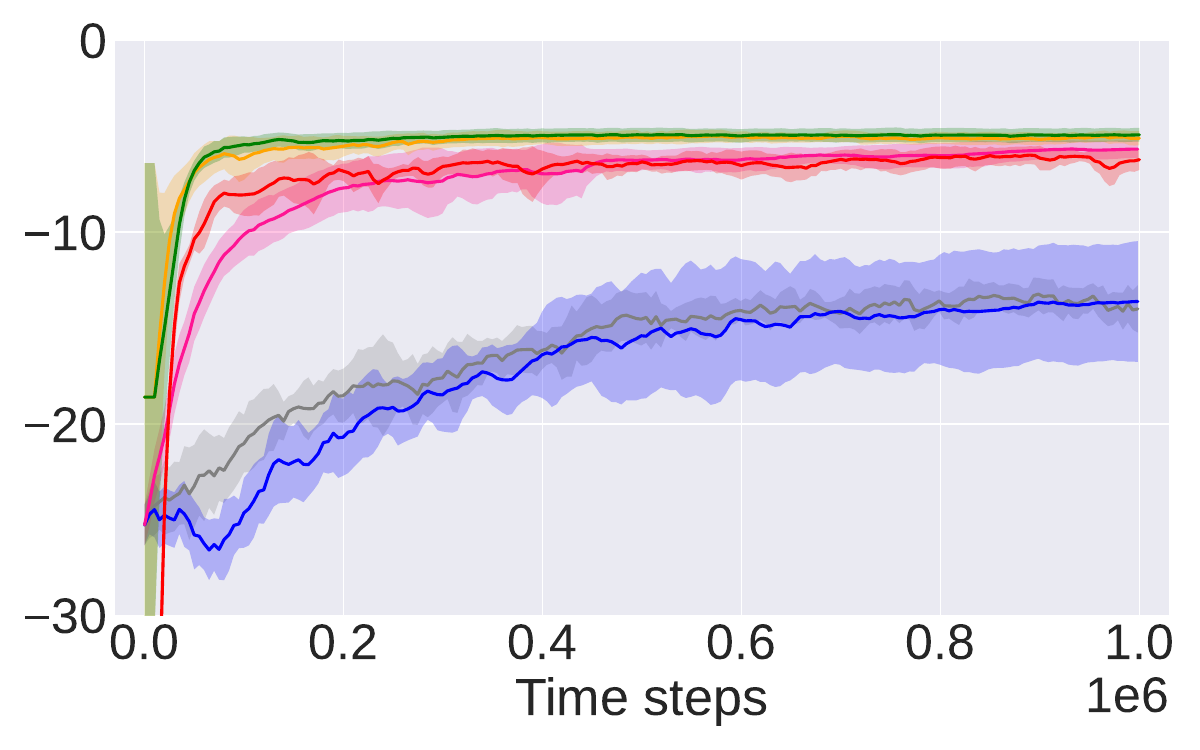}}
	\subfloat[Swimmer]{
		\includegraphics[height=0.20\textwidth]{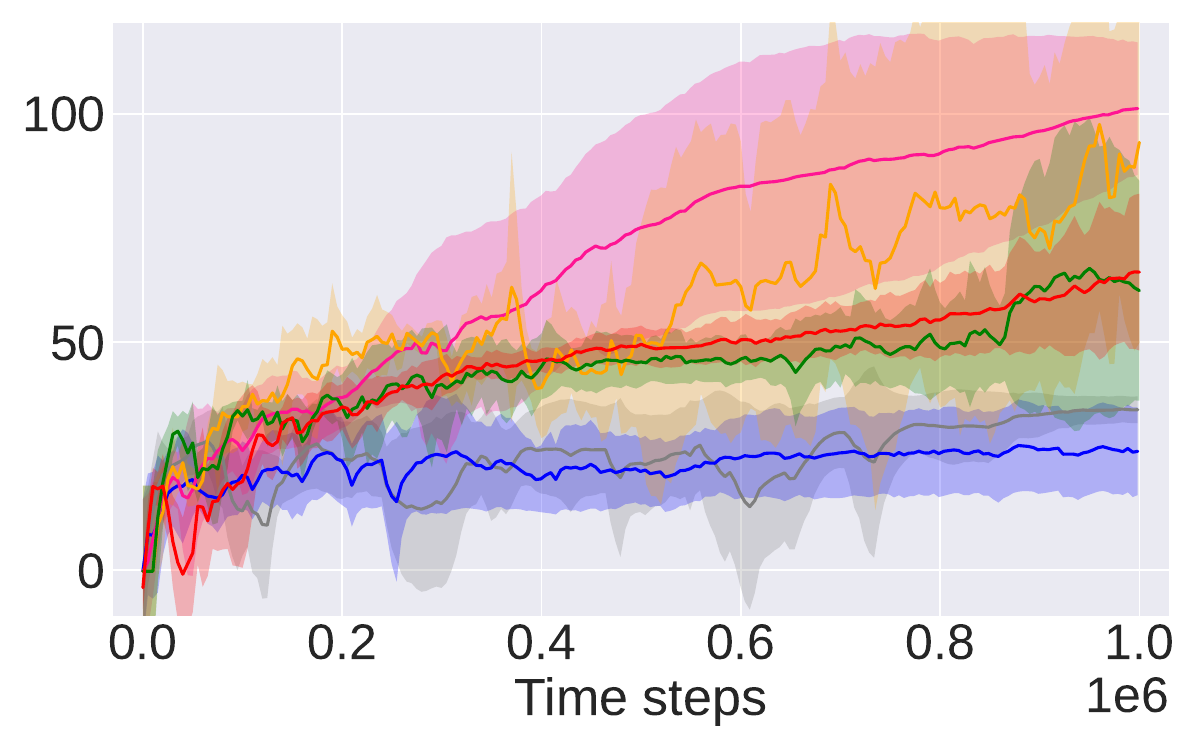}}
	\vfill
	\subfloat[Walker2d]{
		\includegraphics[height=0.20\textwidth, width=0.42\textwidth]{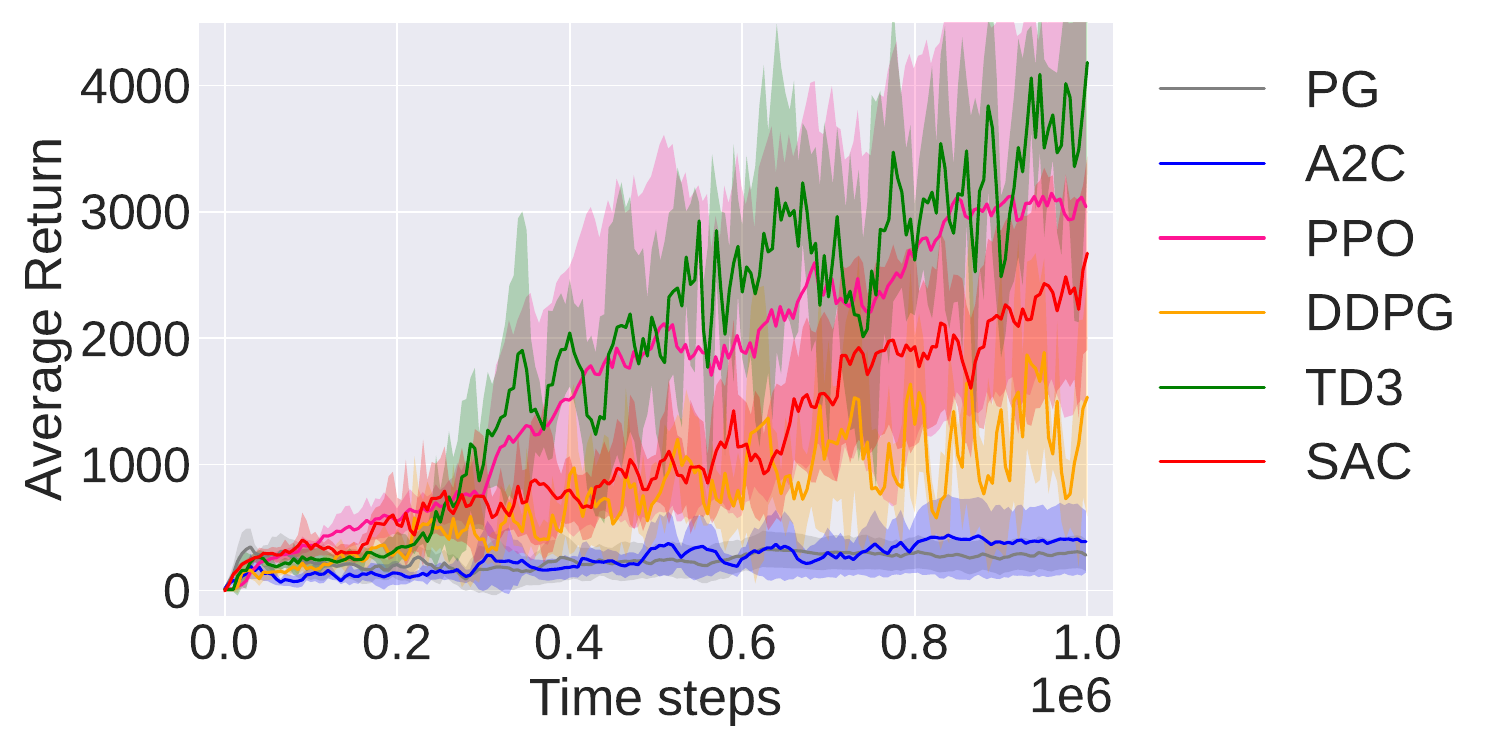}}
	\caption{The learning curves of algorithms in XuanCe on MuJoCo environment tasks.}
	\label{fig:benchmark-mujoco}
\label{fig:benchmark:mujoco}
\end{figure}

Table~\ref{table:benchmark:mujoco:on-policy} and~\ref{table:benchmark:mujoco:off-policy} present the benchmark results for selected on-policy and off-policy algorithms, respectively. 
The published results for A2C and PPO are referenced from \cite{schulman2017proximal}, while the results for DDPG and TD3 are referenced from \cite{fujimoto2018addressing}. Additionally, the published result for SAC is obtained from \cite{haarnoja2018soft}. For a fair comparison, we keep the training parameters of these algorithms consistent with the those in the referenced papers.
From Table~\ref{table:benchmark:mujoco:on-policy} and~\ref{table:benchmark:mujoco:off-policy}, it can be found that, across the 10 continuous control tasks, the algorithms in XuanCe often achieves better results than those published in the original papers.
Figure~\ref{fig:benchmark:mujoco} illustrates the learning curves of the selected algorithms in XuanCe on these control tasks over 5 random seeds, with training steps set to $10^6$. 

% Figure for MuJoCo results.

\subsection{Atari}

The representation capacity of DRL algorithms is important for solving tasks with high dimensional observations. We test DQN and PPO in XuanCe on a set of Atari 2600 environment, which is simulated by Stella and the Arcade Leaning Environment. 
The comparison results over 59 different tasks in Atari 5600 environment are shown in Table~\ref{table:atari}. 

\begin{table}[htbp] \scriptsize
\caption{Benchmark results of Atari for algorithms in XuanCe}
\centering
\begin{tabular}{l|rl|rl|rl|r}
\toprule
               & \multicolumn{4}{c|}{\textbf{DQN}} & \multicolumn{3}{c}{\textbf{PPO}} \\
\midrule
\textbf{Task}  & \multicolumn{2}{c|}{\textbf{XuanCe}} & \multicolumn{2}{c|}{\textbf{Published}} & \multicolumn{2}{c|}{\textbf{XuanCe}} & \textbf{Published} \\
\midrule
AirRaid   	   & 7400.0 & $\pm$2123.38 & - & - & 5865.0 & $\pm$1945.48 & - \\
Alien  	  	   & 1680.0 & $\pm$247.79 & \textbf{3069} & $\pm$1093 & 1056.0 & $\pm$212.85 & \textbf{1850.3} \\
Amidar    	   & 595.33 & $\pm$46.20 & \textbf{739.5} & $\pm$3024 & \textbf{961.0} & $\pm$4.20 & 674.6 \\
Assault   	   & \textbf{10833.33}  & $\pm$2180.16 & 3359 & $\pm$775 & \textbf{6265.67} & $\pm$1884.26 & 4971.9 \\
Asterix   	   & \textbf{9833.33} & $\pm$4401.58 & 6012 & $\pm$1744 & 2590.0 & $\pm$142.83 & \textbf{4532.5} \\
Asteroids 	   & 1246.67 & $\pm$194.82 & \textbf{1629} & $\pm$542 & \textbf{2500.0} & $\pm$373.58 & 2097.5 \\
Atlantis  	   & \textbf{219733.33} & $\pm$151823.17 & 85641 & $\pm$17600 & 939266.67 & $\pm$13112.42 & \textbf{2311815.0} \\
BankHeist  	   & \textbf{1156.67} & $\pm$49.89 & 429.7 & $\pm$650 & 1113.33 & $\pm$4.71 & \textbf{1280.6} \\
BattleZone 	   & \textbf{31800.0} & $\pm$13614.70 & 26300 & $\pm$7725 & 10333.33 & $\pm$5185.45 & \textbf{17366.7} \\
BeamRider  	   & \textbf{7599.6} & $\pm$1371.67 & 6846 & $\pm$1619 & \textbf{1778.67} & $\pm$318.67 & 1590.0 \\
Berzerk    	   & 1130.0 & $\pm$190.53 & - & - & 1750.0 & $\pm$29.44 & - \\
Bowling	   	   & \textbf{78.0} & $\pm$0.0 & 42.4  & $\pm$88 & \textbf{72.8} & $\pm$7.08 & 40.1 \\
Boxing     	   & \textbf{97.33} & $\pm$2.49 & 71.8 & $\pm$8.4 & \textbf{98.67} & $\pm$0.94 & 94.6 \\
Breakout   	   & \textbf{415.33} & $\pm$44.81 & 401.2 & $\pm$26.9 & \textbf{395.33} & $\pm$18.12 & 274.8 \\
Carnival  	   & 2396.0 & $\pm$504.44 & - & - & 5473.33 & $\pm$23.57 & - \\
Centipede  	   & \textbf{9670.0} & $\pm$1522.0 & 8309 & $\pm$5237 & \textbf{4832.33} & $\pm$1932.04 & 4386.4 \\
ChopperCommand & \textbf{19100.0} & $\pm$6125.68 & 6687 & $\pm$2916 & \textbf{8233.33} & $\pm$2524.99 & 3516.3 \\
CrazyClimber   & 96140.0 & $\pm$45187.46 & \textbf{114103} & $\pm$22797 & \textbf{127600.0} & $\pm$9651.25 & 110202.0 \\
Defender       & 8290.0 & $\pm$3281.98 & - & - & 57016.67 & $\pm$3744.4 & - \\
DemonAttack    & \textbf{59328.33} & $\pm$35781.27 & 9711 & $\pm$2406 & 8641.67 & $\pm$1319.26 & \textbf{11378.4} \\
DoubleDunk     & \textbf{1.33} & $\pm$0.94 & -18.1 & $\pm$2.6 & \textbf{0.0} & $\pm$1.63 & -14.9 \\
%ElevatorAction & 66.67 & $\pm$94.28 & - & - & 48566.67 & $\pm$13506.87 & - \\
Enduro  	   & \textbf{2127.0} & $\pm$124.57 & 301.8 & $\pm$24.6 & \textbf{1346.33} & $\pm$244.99 & 758.3 \\
FishingDerby	   & \textbf{25.0} & $\pm$13.02 & -0.8 & $\pm$19.0 & \textbf{35.0} & $\pm$3.27 & 17.8 \\
Freeway   	   & \textbf{32.2} & $\pm$0.98 & 30.3 & $\pm$0.7 & \textbf{33.2} & $\pm$0.4 & 32.5 \\
Frostbite 	   & \textbf{7270.0} & $\pm$1187.94 & 328.3 & $\pm$250.5 & 300.0 & $\pm$14.14 & \textbf{314.2} \\
Gopher 		   & \textbf{11948.0} & $\pm$3763.50 & 8520 & $\pm$3279 & 2626.67 & $\pm$1654.60 & \textbf{2932.9} \\
Gravitar 	   & \textbf{1683.33} & $\pm$589.26 & 306.7 & $\pm$223.9  & 650.0 & $\pm$212.13 & \textbf{737.2} \\
Hero 		   & 17578.0 & $\pm$3386.97 & \textbf{19950} & $\pm$158  & 27543.33 & $\pm$94.63 & - \\
IceHockey 	   & \textbf{0.0} & $\pm$1.26 & -1.6 & $\pm$2.5  & \textbf{-2.0} & $\pm$1.41 & -4.2 \\
Jamesbond 	   & \textbf{2016.67} & $\pm$2498.44 & 576.7 & $\pm$175.5 & \textbf{733.33} & $\pm$117.85 & 560.7 \\
JourneyEscape  & 2800.0 & $\pm$4808.33 & - & -  & -300.0 & $\pm$244.95 & - \\
Kangaroo 	   & 4600.0 &$\pm$1232.88 & \textbf{6740} & $\pm$2959 & 8300.0 & $\pm$1373.56 & \textbf{9928.7} \\
Krull 		   & \textbf{9886.67} & $\pm$669.44 & 3805 & $\pm$1033 & \textbf{9026.67} & $\pm$246.35 & 7942.3 \\
KungFuMaster   & \textbf{53060.0} & $\pm$10103.58 & 23270 & $\pm$5955 & \textbf{41666.68} & $\pm$2317.09 & 23310.3 \\
MontezumaRevenge & 0.0 & $\pm$0 & 0.0 & $\pm$0 & \textbf{66.67} & $\pm$47.14 & 42.0 \\
MsPacman  	     & \textbf{3896.0} & $\pm$427.06 & 2311 & $\pm$525 & \textbf{3388.0} & $\pm$939.16 & 2096.5 \\
NameThisGame 	 & 7124.0 & $\pm$3091.93 & \textbf{7257} & $\pm$547 & \textbf{11553.33} & $\pm$1610.43 & 6254.9 \\
Phoenix 		 & 11550 & $\pm$284.37 & - & - & 16610 & $\pm$5847.79 & - \\
Pitfall 		 & 0.0 & $\pm$0.0 & -  & - & \textbf{-24.2} & $\pm$38.76 & -32.9 \\
Pong 	  	   	 & 17.67 & $\pm$1.23 & \textbf{18.9} & $\pm$1.3 & 19.67 & $\pm$0.94 & \textbf{20.7} \\
Pooyan 		 	 & 8022.0 & $\pm$2508.53 & - & - & 6245.0 & $\pm$1562.97 & - \\
PrivateEye 		 & 966.67 & $\pm$2176.13 & \textbf{1788} & $\pm$5473 & 30.0 & $\pm$45.83 & \textbf{69.5} \\
Qbert	  	   	 & \textbf{12585.0} & $\pm$2528.86 & 10596 & $\pm$3294 & \textbf{19085.0} & $\pm$2410.43 & 14293.3 \\
Riverraid 		 & \textbf{13480.0} & $\pm$576.02 & 8316 & $\pm$1049 & \textbf{13053.33} & $\pm$765.26 & 8393.6 \\
RoadRunner 		 & \textbf{59000.0} & $\pm$3894.44 & 18257 & $\pm$4268 & \textbf{40633.33} & $\pm$8491.7 & 25076.0 \\
Robotank 		 & \textbf{66.33} & $\pm$10.78 & 51.6 & $\pm$4.7 & \textbf{15.0} & $\pm$1.63 & 5.5 \\
Seaquest 		 & \textbf{5400.0} & $\pm$762.59 & 5286 & $\pm$1310 & \textbf{1720.0} & $\pm$56.57 & 1204.5 \\
Skiing 			 & -12313.33 & $\pm$766.81 & -  & - & -15710.0 & $\pm$9.63 & - \\
Solaris 		 & 3600.0 & $\pm$2673.33 & - & - & 9260.0 & $\pm$3572.23 & - \\
SpaceInvaders 	 & 1630.0 & $\pm$375.73 & \textbf{1976} & $\pm$893 & 908.33 & $\pm$243.05 & \textbf{942.5} \\
StarGunner 		 & 37280.0 & $\pm$12316.88 & \textbf{57997} & $\pm$3152 & 27566.67 & $\pm$3564.95 & \textbf{32689.0} \\
Tennis 			 & \textbf{1.0} & $\pm$0.82 & -2.5 & $\pm$1.9 & \textbf{-1.33} & $\pm$0.47 & -14.8 \\
TimePilot 		 & \textbf{8260.0} & $\pm$338.23 & 5947 & $\pm$1600 & \textbf{6800.0} & $\pm$1920.07  & 4342.0 \\
Tutankham 		 & \textbf{214.0} & $\pm$15.56 & 186.7 & $\pm$41.9 & 174.0 & $\pm$5.89 & \textbf{254.4} \\
UpNDown 		 & \textbf{22923.33} & $\pm$1249.22 & 8456 & $\pm$3162 & \textbf{282200.0} & $\pm$4199.53 & 95445.0 \\
Venture 		 & \textbf{560.0} & $\pm$174.36 & 380.0 & $\pm$238.6 & \textbf{66.67} & $\pm$94.28 & 0.0 \\
VideoPinball 	 & \textbf{847303.67} & $\pm$179889.24 & 42684 & $\pm$16287 & \textbf{100107.66} & $\pm$49125.29  & 37389.0 \\
WizardOfWor 		 & 1560.0 & $\pm$120.0 & \textbf{3393} & $\pm$2019 & 1766.67 & $\pm$339.93 & \textbf{4185.3} \\
Zaxxon 			 & 4820.0 & $\pm$2429.32 & \textbf{4977} & $\pm$1235 & \textbf{5866.67} & $\pm$1087.3 & 5008.7 \\
\midrule
\# Higher  & 35 & & 13 & & 33 & & 16 \\
\bottomrule
\end{tabular}
\label{table:atari}
\end{table}

In Table~\ref{table:atari}, the published results of DQN and PPO are obtained from \cite{mnih2015human} and \cite{schulman2017proximal}, respectively. We keep the consistency in the neural network structure and hyper-parameters with those in the original papers. The CNN-based representation modules are selected for DQN and PPO to extract features from the original images. 
For DQN algorithm, we run five environments in parallel to accelerate the experience sampling and set the number of total training steps to $5\times 10^7$. Considering that the on-policy algorithm PPO does not significantly consume memory for the experience replay buffer, we run eight parallel environments to enable agent to interact with the environments more efficiently. The number of total training steps for PPO is set to $10^7$. We select the highest score across all intermediate evaluation phases. In each evaluation phase, we test the model five times and calculate the average accumulated rewards as the current score. It could be found that XuanCe demonstrates strong performance in Atari tasks when compared with the results in the published papers.

\subsection{StarCraftII Multi-Agent Challenge}

The SMAC is a widely-used environment for cooperative multi-agent reinforcement learning based on Blizzard's StarCraft II RTS game~\citep{vinyals2017starcraft}. It includes multiple maps with varying levels of difficulty.
In this environment, the agents aim to eliminate all enemies on the opposing side through cooperation to win the game. Each agent observes partially of the global state, that brings uncertainty for agent to make decisions. Hence, we choose RNN-based representations that leverages the historical information of the sequence data to alleviate the issue of incomplete information caused by partial observability. The recurrent layer in each RNN-based representation module is implemented using a grated recurrent unit (GRU) with 64-dimentional hidden states. 
In this section, nine different maps are selected to evaluate the performance of MARL algorithms in XuanCe. Detailed information about these maps for evaluation is presented in Table~\ref{table:benchmark:smac:maps}.

\begin{table}[htbp]\footnotesize
\caption{The features of the selected maps in SMAC environment}
\centering
\begin{tabular}{l|ccccccccc}
\toprule
\textbf{Map} & 2m\_vs\_1z & 3m & 8m & 1c3s5z & 2s3z & 25m & 5m\_vs\_6m & 8m\_vs\_9m & MMM2 \\
\midrule
n\_agents &	2	&	3	&	8	&	9	&	5	&	25	&	5	&	8	&	10	\\
n\_enemies &	1 &	3 &	8 &	9 &	5	&	25	&	6	&	9	&	12	\\
episode\_steps	&	150	&	60	&	120	&	180	&	120	&	150	&	70	&	120	&	180	\\
training\_steps	&	1M	&	1M	&	1M	&	2M	&	2M	&	5M	&	10M	&	10M	&	10M	\\
\bottomrule
\end{tabular}
\label{table:benchmark:smac:maps}
\end{table}

We test seven typical MARL algorithms in XuanCe on the selected maps of SMAC environment, including IQL~\citep{sunehag2017value}, VDN~\citep{sunehag2017value}, QMIX~\citep{rashid2020monotonic}, WQMIX~\citep{rashid2020weighted}, VDAC-mix~\citep{su2021value}, IPPO~\citep{yu2022surprising}, and MAPPO~\citep{yu2022surprising}.
The IQL, VDN, QMIX, and WQMIX are off-policy and value-based MARL algorithms, while the VDAC-mix, IPPO, and MAPPO are on-policy and policy-based MARL algorithms. 
We use the parameter sharing trick for all agents to reduce the complexity of the models and improve the training speed. Additionally, to improve sample efficiency, we create eight environments and run them in parallel for interaction. 
The hyper-parameters and network structures of these algorithms are kept consistent with those specified in the original papers. 

\begin{figure}[h]
	\centering
	\subfloat[2m\_vs\_1z]{
		\includegraphics[height=0.20\textwidth]{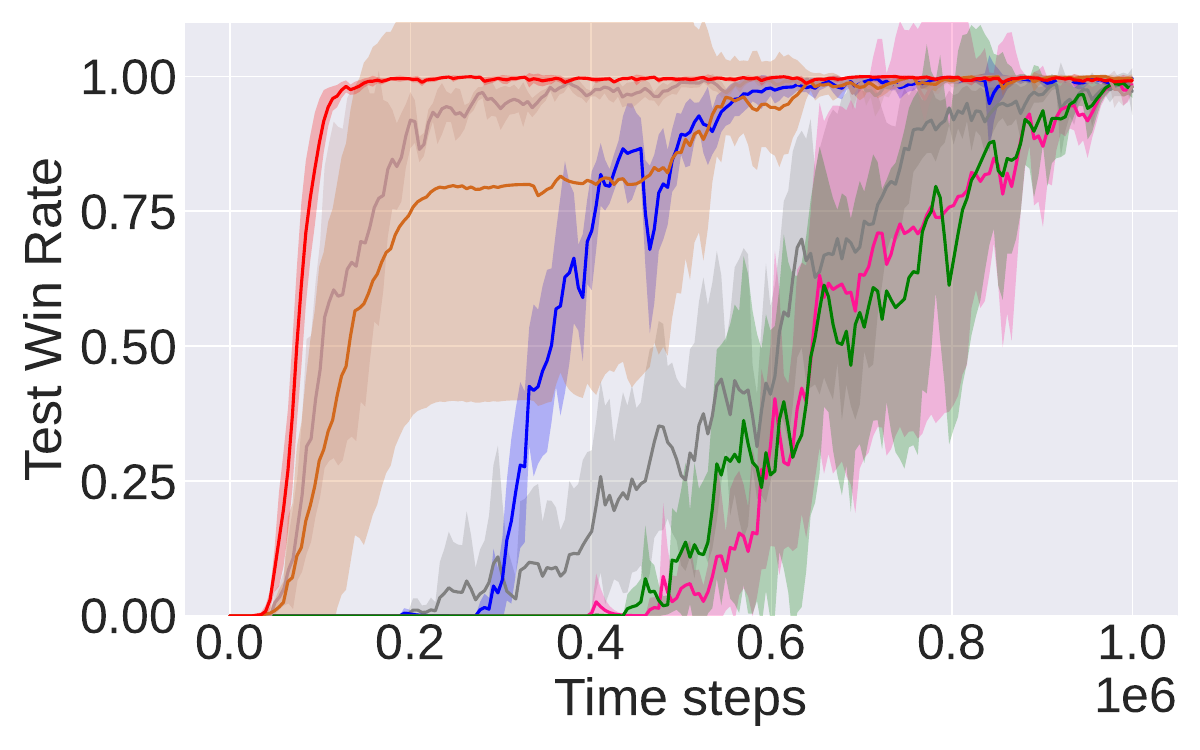}}
	\subfloat[3m]{
		\includegraphics[height=0.20\textwidth]{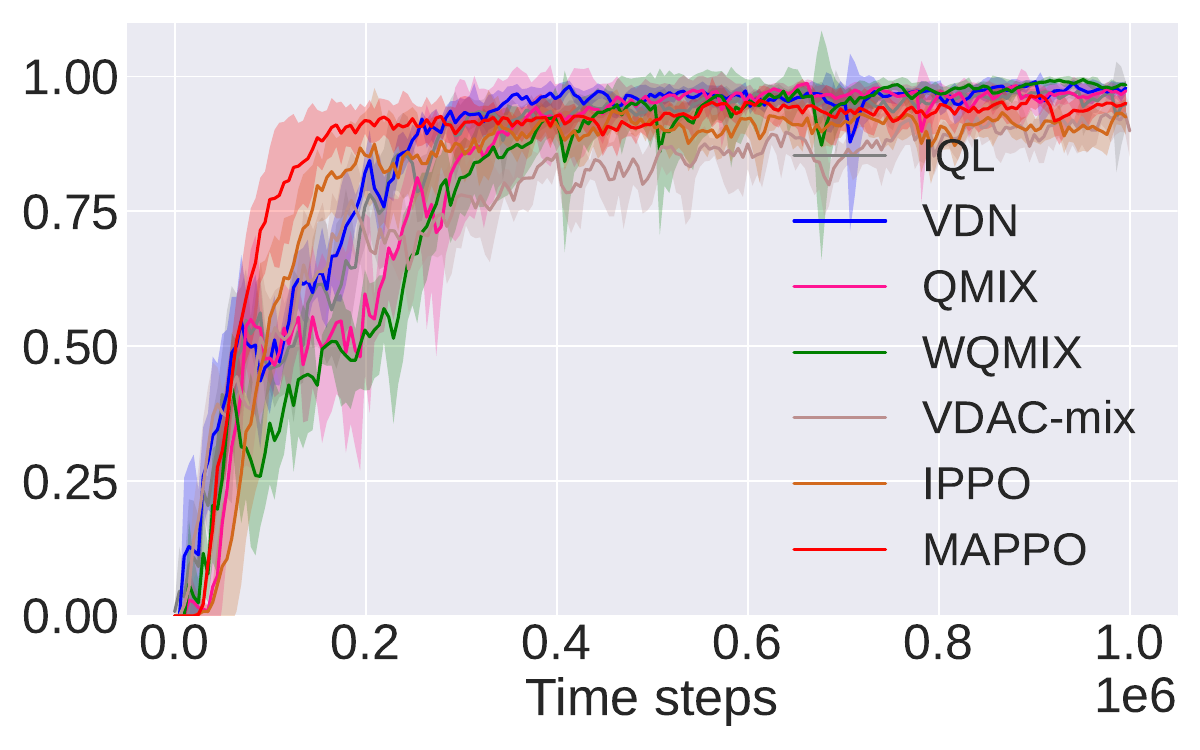}}
	\subfloat[8m]{
		\includegraphics[height=0.20\textwidth]{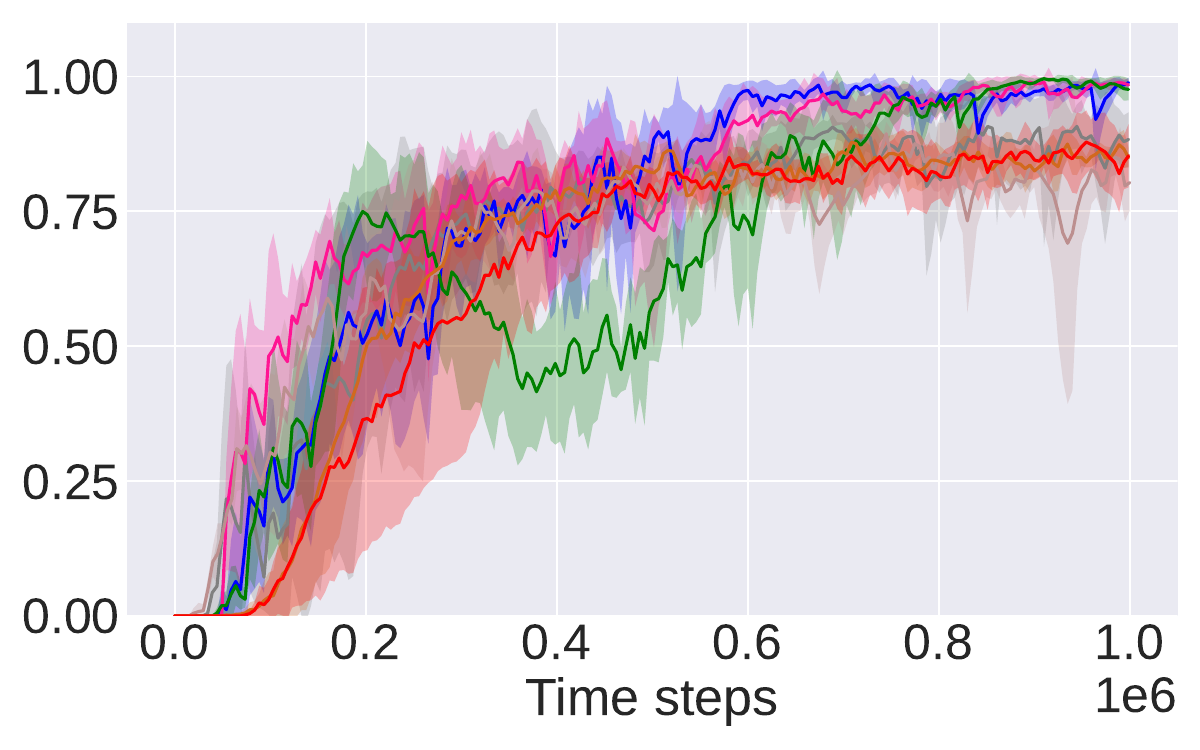}}
	\vfill
	\subfloat[1c3s5z]{
		\includegraphics[height=0.20\textwidth]{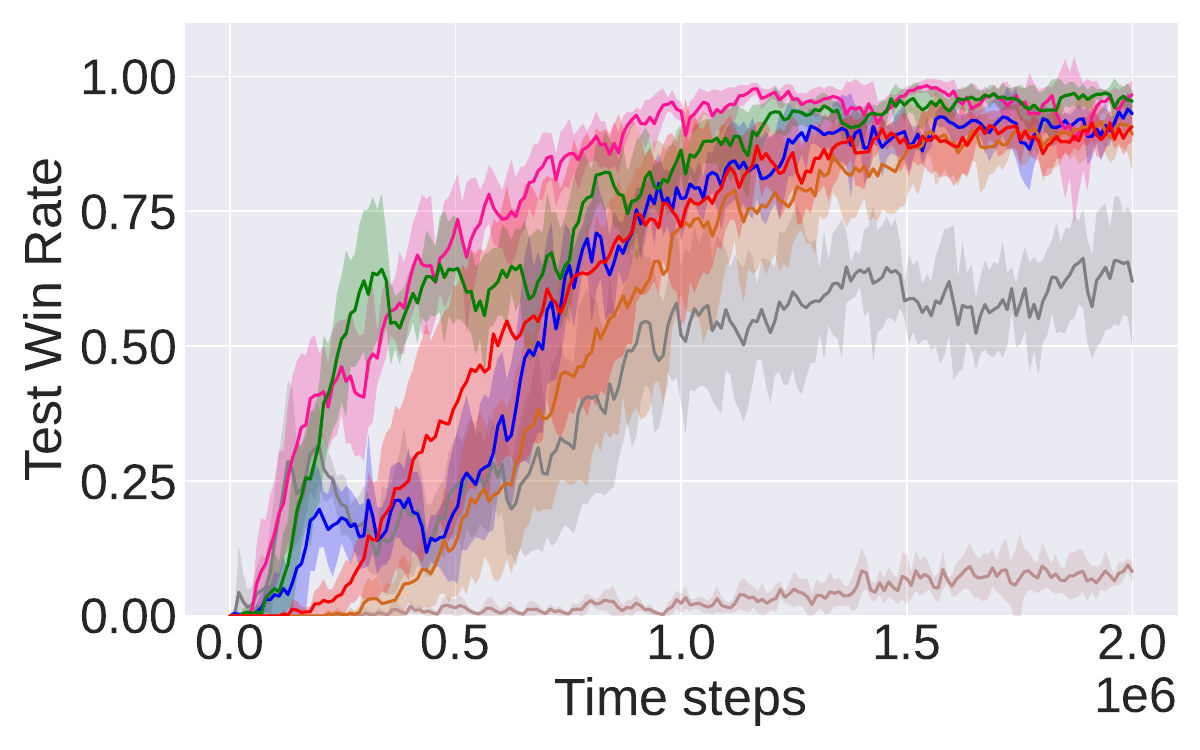}}
	\subfloat[2s3z]{
		\includegraphics[height=0.20\textwidth]{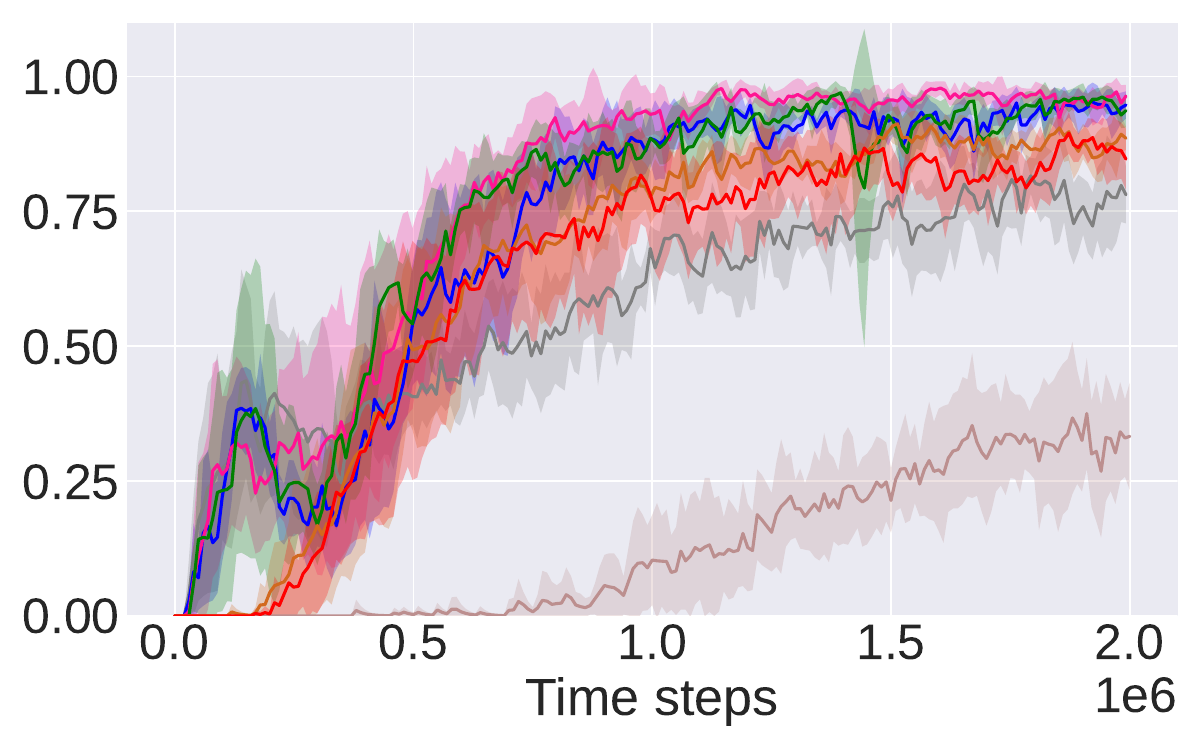}}
	\subfloat[25m]{
		\includegraphics[height=0.20\textwidth]{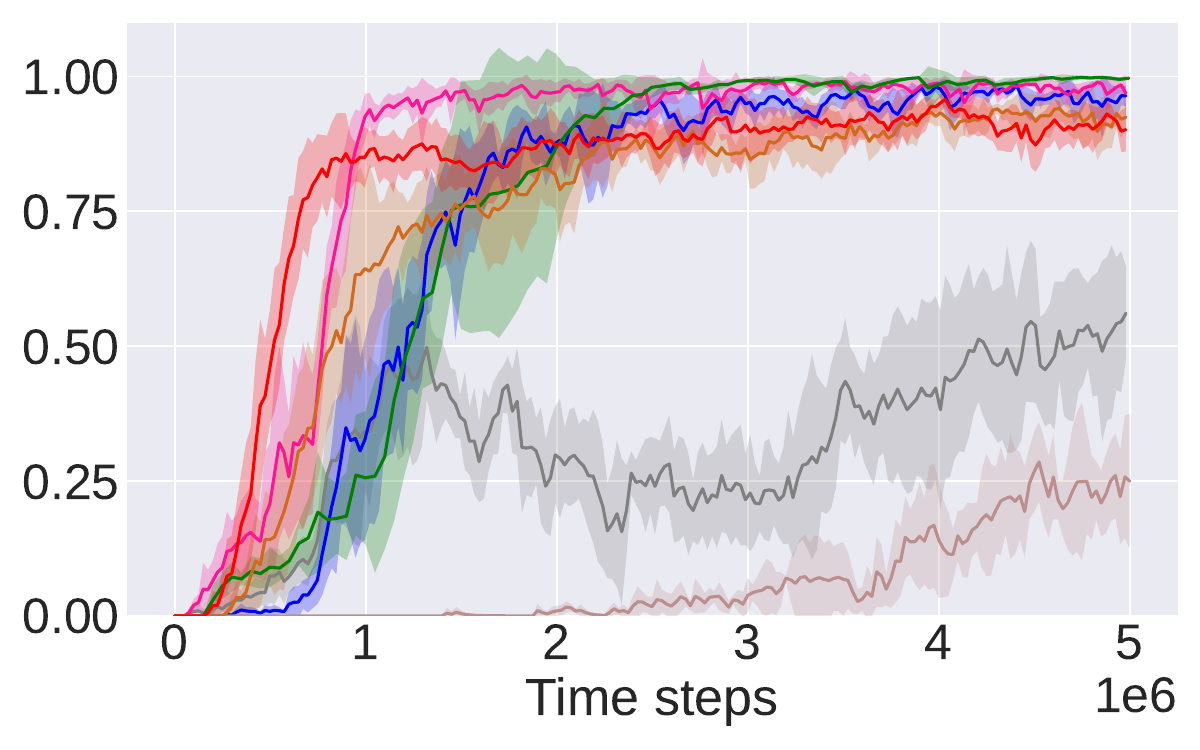}}
	\vfill
	\subfloat[5m\_vs\_6m]{
		\includegraphics[height=0.20\textwidth]{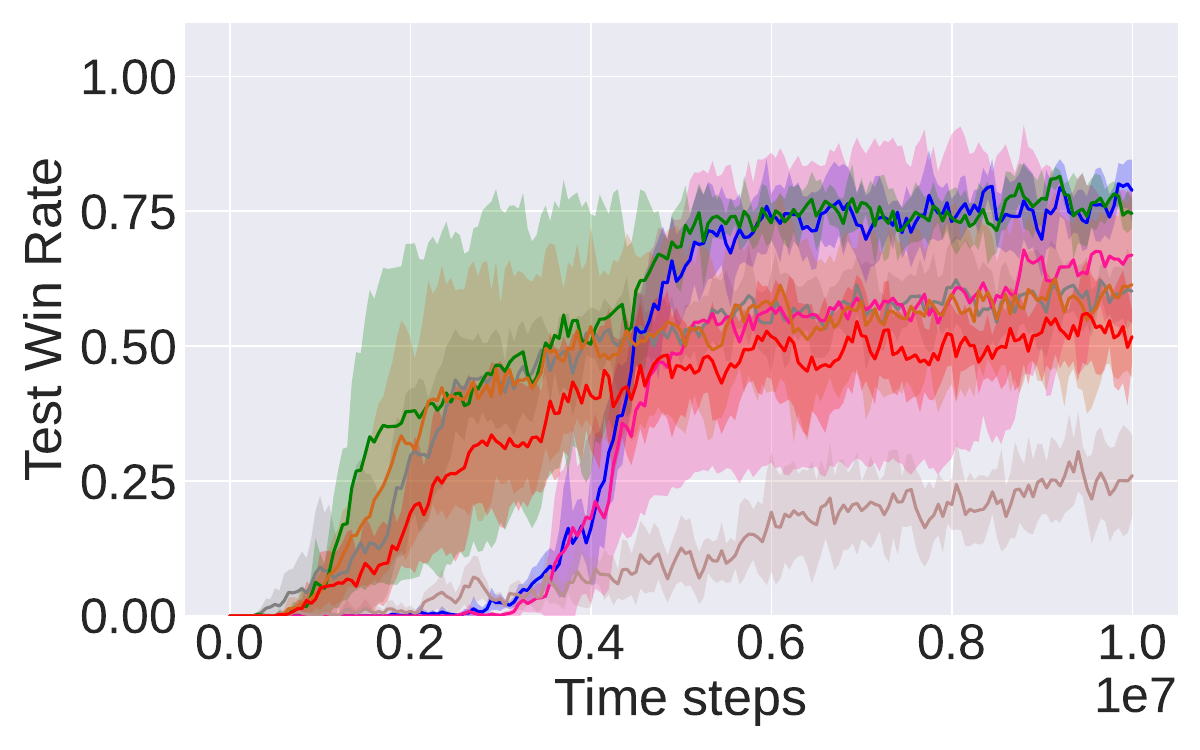}}
	\subfloat[8m\_vs\_9m]{
		\includegraphics[height=0.20\textwidth]{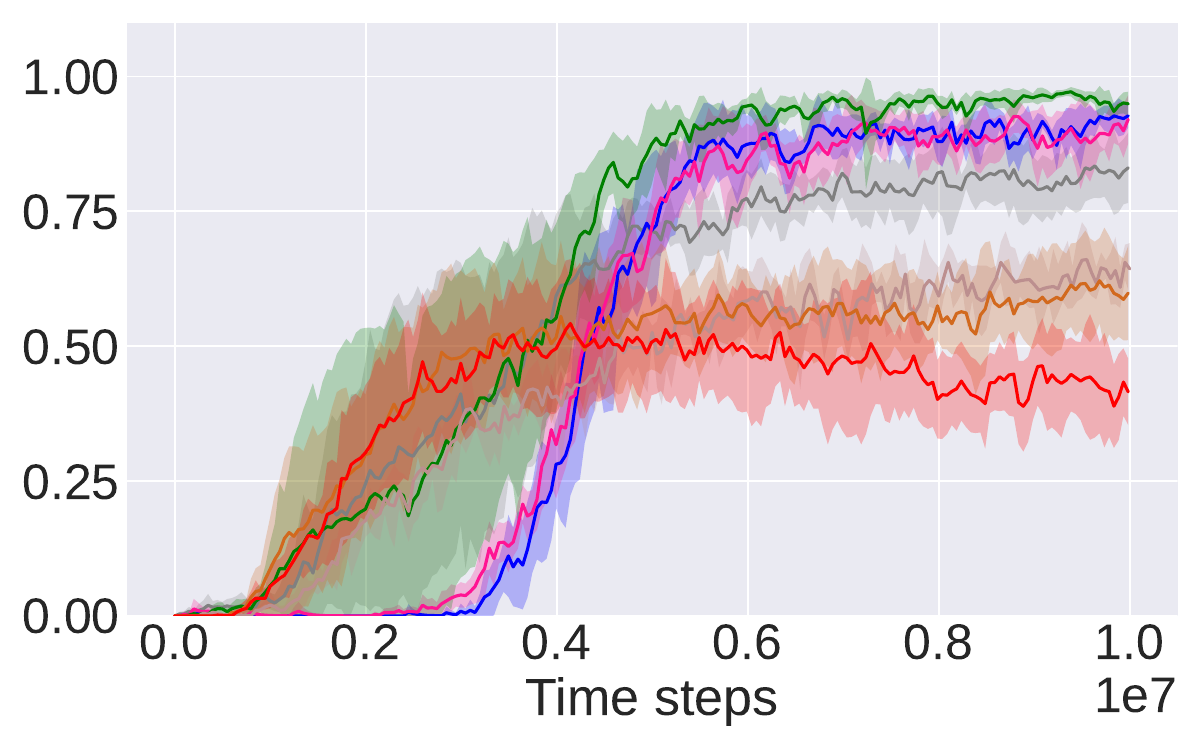}}
	\subfloat[MMM2]{
		\includegraphics[height=0.20\textwidth]{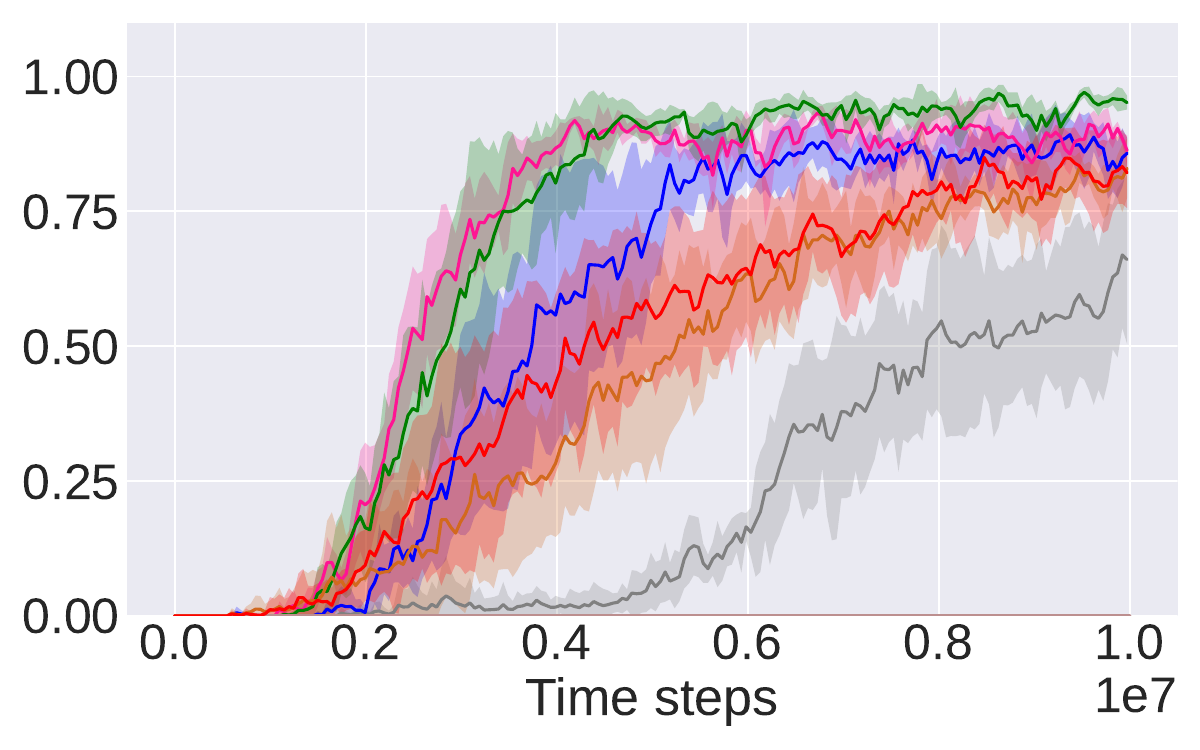}}
	\caption{The learning curves of algorithms in XuanCe on SMAC environment maps.}
	\label{fig:benchmark-smac}
\end{figure}

Figure~\ref{fig:benchmark-smac} illustrates the learning curves of the selected MARL algorithms on the nine maps across five random seeds. It shows the increase in win rate for each algorithm during the training process. The win rate is achieved by calculating the ratio of win times to the total testing episodes during each intermediate evaluation phase. Figure~\ref{fig:benchmark-smac} demonstrates that the MARL algorithms in this library could exhibit good learning performance on SMAC environment. 

\section{Documentation}

The full documentation of XuanCe can be achieved at \href{https://xuance.readthedocs.io/}{https://xuance.readthedocs.io/}. It is written by reStructuredText, a popular file format for technical documentation of Python projects. Recently, we have provided documentation in two languages: English and Chinese. The documentation includes technical details about the library's installation, usage, APIs, and benchmark results, etc. 
It also provides examples of typical algorithms on various tasks, which can be a reference for users when implementing their own tasks. Besides, it introduces the method of extending XuanCe with new environments and algorithms, enabling users to contribute new implementations.

% Please add the following required packages to your document preamble:
% \usepackage{multirow}

\bibliography{references.bib}

\begin{thebibliography}{32}
\providecommand{\natexlab}[1]{#1}
\providecommand{\url}[1]{\texttt{#1}}
\expandafter\ifx\csname urlstyle\endcsname\relax
  \providecommand{\doi}[1]{doi: #1}\else
  \providecommand{\doi}{doi: \begingroup \urlstyle{rm}\Url}\fi

\bibitem[Abadi et~al.(2016)Abadi, Barham, Chen, Chen, Davis, Dean, Devin,
  Ghemawat, Irving, Isard, et~al.]{abadi2016tensorflow}
Mart{\'\i}n Abadi, Paul Barham, Jianmin Chen, Zhifeng Chen, Andy Davis, Jeffrey
  Dean, Matthieu Devin, Sanjay Ghemawat, Geoffrey Irving, Michael Isard, et~al.
\newblock Tensorflow: a system for large-scale machine learning.
\newblock In \emph{Osdi}, volume~16, pages 265--283. Savannah, GA, USA, 2016.

\bibitem[Achiam(2018)]{SpinningUp2018}
Joshua Achiam.
\newblock {Spinning Up in Deep Reinforcement Learning}.
\newblock 2018.

\bibitem[Brockman et~al.(2016)Brockman, Cheung, Pettersson, Schneider,
  Schulman, Tang, and Zaremba]{brockman2016openai}
Greg Brockman, Vicki Cheung, Ludwig Pettersson, Jonas Schneider, John Schulman,
  Jie Tang, and Wojciech Zaremba.
\newblock Openai gym.
\newblock \emph{arXiv preprint arXiv:1606.01540}, 2016.

\bibitem[Castro et~al.(2018)Castro, Moitra, Gelada, Kumar, and
  Bellemare]{castro18dopamine}
Pablo~Samuel Castro, Subhodeep Moitra, Carles Gelada, Saurabh Kumar, and
  Marc~G. Bellemare.
\newblock Dopamine: {A} {R}esearch {F}ramework for {D}eep {R}einforcement
  {L}earning.
\newblock 2018.
\newblock URL \url{http://arxiv.org/abs/1812.06110}.

\bibitem[D'Eramo et~al.(2021)D'Eramo, Tateo, Bonarini, Restelli, and
  Peters]{JMLR:v22:18-056}
Carlo D'Eramo, Davide Tateo, Andrea Bonarini, Marcello Restelli, and Jan
  Peters.
\newblock Mushroomrl: Simplifying reinforcement learning research.
\newblock \emph{Journal of Machine Learning Research}, 22\penalty0
  (131):\penalty0 1--5, 2021.
\newblock URL \url{http://jmlr.org/papers/v22/18-056.html}.

\bibitem[Ding et~al.(2020)Ding, Yu, Huang, Zhang, Mai, and Dong]{ding2020rlzoo}
Zihan Ding, Tianyang Yu, Yanhua Huang, Hongming Zhang, Luo Mai, and Hao Dong.
\newblock Rlzoo: A comprehensive and adaptive reinforcement learning library.
\newblock \emph{arXiv preprint arXiv:2009.08644}, 2020.

\bibitem[Fujimoto et~al.(2018)Fujimoto, Hoof, and
  Meger]{fujimoto2018addressing}
Scott Fujimoto, Herke Hoof, and David Meger.
\newblock Addressing function approximation error in actor-critic methods.
\newblock In \emph{International conference on machine learning}, pages
  1587--1596. PMLR, 2018.

\bibitem[Fujita et~al.(2021)Fujita, Nagarajan, Kataoka, and
  Ishikawa]{fujita2021chainerrl}
Yasuhiro Fujita, Prabhat Nagarajan, Toshiki Kataoka, and Takahiro Ishikawa.
\newblock Chainerrl: A deep reinforcement learning library.
\newblock \emph{The Journal of Machine Learning Research}, 22\penalty0
  (1):\penalty0 3557--3570, 2021.

\bibitem[Haarnoja et~al.(2018)Haarnoja, Zhou, Abbeel, and
  Levine]{haarnoja2018soft}
Tuomas Haarnoja, Aurick Zhou, Pieter Abbeel, and Sergey Levine.
\newblock Soft actor-critic: Off-policy maximum entropy deep reinforcement
  learning with a stochastic actor.
\newblock In \emph{International conference on machine learning}, pages
  1861--1870. PMLR, 2018.

\bibitem[Hu et~al.(2023)Hu, Zhong, Gao, Wang, Dong, Liang, Li, Chang, and
  Yang]{hu2023marllib}
Siyi Hu, Yifan Zhong, Minquan Gao, Weixun Wang, Hao Dong, Xiaodan Liang, Zhihui
  Li, Xiaojun Chang, and Yaodong Yang.
\newblock Marllib: A scalable and efficient library for multi-agent
  reinforcement learning.
\newblock \emph{Journal of Machine Learning Research}, 24:\penalty0 1--23,
  2023.

\bibitem[Huawei Technologies~Co.(2022)]{huawei2022huawei}
Ltd. Huawei Technologies~Co.
\newblock Huawei mindspore ai development framework.
\newblock In \emph{Artificial Intelligence Technology}, pages 137--162.
  Springer, 2022.

\bibitem[Hubbs et~al.(2020)Hubbs, Perez, Sarwar, Sahinidis, Grossmann, and
  Wassick]{hubbs2020or}
Christian~D Hubbs, Hector~D Perez, Owais Sarwar, Nikolaos~V Sahinidis,
  Ignacio~E Grossmann, and John~M Wassick.
\newblock Or-gym: A reinforcement learning library for operations research
  problems.
\newblock \emph{arXiv preprint arXiv:2008.06319}, 2020.

\bibitem[Kurach et~al.(2020)Kurach, Raichuk, Sta{\'n}czyk, Zaj{\k{a}}c, Bachem,
  Espeholt, Riquelme, Vincent, Michalski, Bousquet, et~al.]{kurach2020google}
Karol Kurach, Anton Raichuk, Piotr Sta{\'n}czyk, Micha{\l} Zaj{\k{a}}c, Olivier
  Bachem, Lasse Espeholt, Carlos Riquelme, Damien Vincent, Marcin Michalski,
  Olivier Bousquet, et~al.
\newblock Google research football: A novel reinforcement learning environment.
\newblock In \emph{Proceedings of the AAAI conference on artificial
  intelligence}, volume~34, pages 4501--4510, 2020.

\bibitem[LeCun et~al.(2015)LeCun, Bengio, and Hinton]{lecun2015deep}
Yann LeCun, Yoshua Bengio, and Geoffrey Hinton.
\newblock Deep learning.
\newblock \emph{nature}, 521\penalty0 (7553):\penalty0 436--444, 2015.

\bibitem[Liang et~al.(2018)Liang, Liaw, Nishihara, Moritz, Fox, Goldberg,
  Gonzalez, Jordan, and Stoica]{liang2018rllib}
Eric Liang, Richard Liaw, Robert Nishihara, Philipp Moritz, Roy Fox, Ken
  Goldberg, Joseph Gonzalez, Michael Jordan, and Ion Stoica.
\newblock Rllib: Abstractions for distributed reinforcement learning.
\newblock In \emph{International Conference on Machine Learning}, pages
  3053--3062. PMLR, 2018.

\bibitem[Liu et~al.(2021)Liu, Yang, Gao, and Wang]{liu2021finrl}
Xiao-Yang Liu, Hongyang Yang, Jiechao Gao, and Christina~Dan Wang.
\newblock Finrl: Deep reinforcement learning framework to automate trading in
  quantitative finance.
\newblock In \emph{Proceedings of the Second ACM International Conference on AI
  in Finance}, pages 1--9, 2021.

\bibitem[Mnih et~al.(2015)Mnih, Kavukcuoglu, Silver, Rusu, Veness, Bellemare,
  Graves, Riedmiller, Fidjeland, Ostrovski, et~al.]{mnih2015human}
Volodymyr Mnih, Koray Kavukcuoglu, David Silver, Andrei~A Rusu, Joel Veness,
  Marc~G Bellemare, Alex Graves, Martin Riedmiller, Andreas~K Fidjeland, Georg
  Ostrovski, et~al.
\newblock Human-level control through deep reinforcement learning.
\newblock \emph{nature}, 518\penalty0 (7540):\penalty0 529--533, 2015.

\bibitem[Pardo(2020)]{pardo2020tonic}
Fabio Pardo.
\newblock Tonic: A deep reinforcement learning library for fast prototyping and
  benchmarking.
\newblock \emph{arXiv preprint arXiv:2011.07537}, 2020.

\bibitem[Paszke et~al.(2019)Paszke, Gross, Massa, Lerer, Bradbury, Chanan,
  Killeen, Lin, Gimelshein, Antiga, et~al.]{paszke2019pytorch}
Adam Paszke, Sam Gross, Francisco Massa, Adam Lerer, James Bradbury, Gregory
  Chanan, Trevor Killeen, Zeming Lin, Natalia Gimelshein, Luca Antiga, et~al.
\newblock Pytorch: An imperative style, high-performance deep learning library.
\newblock \emph{Advances in neural information processing systems}, 32, 2019.

\bibitem[Radford et~al.(2019)Radford, Wu, Child, Luan, Amodei, Sutskever,
  et~al.]{radford2019language}
Alec Radford, Jeffrey Wu, Rewon Child, David Luan, Dario Amodei, Ilya
  Sutskever, et~al.
\newblock Language models are unsupervised multitask learners.
\newblock \emph{OpenAI blog}, 1\penalty0 (8):\penalty0 9, 2019.

\bibitem[Rashid et~al.(2020{\natexlab{a}})Rashid, Farquhar, Peng, and
  Whiteson]{rashid2020weighted}
Tabish Rashid, Gregory Farquhar, Bei Peng, and Shimon Whiteson.
\newblock Weighted qmix: Expanding monotonic value function factorisation for
  deep multi-agent reinforcement learning.
\newblock \emph{Advances in neural information processing systems},
  33:\penalty0 10199--10210, 2020{\natexlab{a}}.

\bibitem[Rashid et~al.(2020{\natexlab{b}})Rashid, Samvelyan, De~Witt, Farquhar,
  Foerster, and Whiteson]{rashid2020monotonic}
Tabish Rashid, Mikayel Samvelyan, Christian~Schroeder De~Witt, Gregory
  Farquhar, Jakob Foerster, and Shimon Whiteson.
\newblock Monotonic value function factorisation for deep multi-agent
  reinforcement learning.
\newblock \emph{The Journal of Machine Learning Research}, 21\penalty0
  (1):\penalty0 7234--7284, 2020{\natexlab{b}}.

\bibitem[Schulman et~al.(2017)Schulman, Wolski, Dhariwal, Radford, and
  Klimov]{schulman2017proximal}
John Schulman, Filip Wolski, Prafulla Dhariwal, Alec Radford, and Oleg Klimov.
\newblock Proximal policy optimization algorithms.
\newblock \emph{arXiv preprint arXiv:1707.06347}, 2017.

\bibitem[Seno and Imai(2022)]{seno2022d3rlpy}
Takuma Seno and Michita Imai.
\newblock d3rlpy: An offline deep reinforcement learning library.
\newblock \emph{The Journal of Machine Learning Research}, 23\penalty0
  (1):\penalty0 14205--14224, 2022.

\bibitem[Serrano-Munoz et~al.(2023)Serrano-Munoz, Chrysostomou, B{\o}gh, and
  Arana-Arexolaleiba]{serrano2023skrl}
Antonio Serrano-Munoz, Dimitrios Chrysostomou, Simon B{\o}gh, and Nestor
  Arana-Arexolaleiba.
\newblock skrl: Modular and flexible library for reinforcement learning.
\newblock \emph{Journal of Machine Learning Research}, 24\penalty0
  (254):\penalty0 1--9, 2023.

\bibitem[Silver et~al.(2018)Silver, Hubert, Schrittwieser, Antonoglou, Lai,
  Guez, Lanctot, Sifre, Kumaran, Graepel, et~al.]{silver2018general}
David Silver, Thomas Hubert, Julian Schrittwieser, Ioannis Antonoglou, Matthew
  Lai, Arthur Guez, Marc Lanctot, Laurent Sifre, Dharshan Kumaran, Thore
  Graepel, et~al.
\newblock A general reinforcement learning algorithm that masters chess, shogi,
  and go through self-play.
\newblock \emph{Science}, 362\penalty0 (6419):\penalty0 1140--1144, 2018.

\bibitem[Su et~al.(2021)Su, Adams, and Beling]{su2021value}
Jianyu Su, Stephen Adams, and Peter Beling.
\newblock Value-decomposition multi-agent actor-critics.
\newblock In \emph{Proceedings of the AAAI conference on artificial
  intelligence}, volume~35, pages 11352--11360, 2021.

\bibitem[Sunehag et~al.(2017)Sunehag, Lever, Gruslys, Czarnecki, Zambaldi,
  Jaderberg, Lanctot, Sonnerat, Leibo, Tuyls, et~al.]{sunehag2017value}
Peter Sunehag, Guy Lever, Audrunas Gruslys, Wojciech~Marian Czarnecki, Vinicius
  Zambaldi, Max Jaderberg, Marc Lanctot, Nicolas Sonnerat, Joel~Z Leibo, Karl
  Tuyls, et~al.
\newblock Value-decomposition networks for cooperative multi-agent learning.
\newblock \emph{arXiv preprint arXiv:1706.05296}, 2017.

\bibitem[Vinyals et~al.(2017)Vinyals, Ewalds, Bartunov, Georgiev, Vezhnevets,
  Yeo, Makhzani, K{\"u}ttler, Agapiou, Schrittwieser,
  et~al.]{vinyals2017starcraft}
Oriol Vinyals, Timo Ewalds, Sergey Bartunov, Petko Georgiev, Alexander~Sasha
  Vezhnevets, Michelle Yeo, Alireza Makhzani, Heinrich K{\"u}ttler, John
  Agapiou, Julian Schrittwieser, et~al.
\newblock Starcraft ii: A new challenge for reinforcement learning.
\newblock \emph{arXiv preprint arXiv:1708.04782}, 2017.

\bibitem[Weng et~al.(2022)Weng, Chen, Yan, You, Duburcq, Zhang, Su, Su, and
  Zhu]{weng2022tianshou}
Jiayi Weng, Huayu Chen, Dong Yan, Kaichao You, Alexis Duburcq, Minghao Zhang,
  Yi~Su, Hang Su, and Jun Zhu.
\newblock Tianshou: A highly modularized deep reinforcement learning library.
\newblock \emph{Journal of Machine Learning Research}, 23\penalty0
  (267):\penalty0 1--6, 2022.

\bibitem[Yu et~al.(2022)Yu, Velu, Vinitsky, Gao, Wang, Bayen, and
  Wu]{yu2022surprising}
Chao Yu, Akash Velu, Eugene Vinitsky, Jiaxuan Gao, Yu~Wang, Alexandre Bayen,
  and Yi~Wu.
\newblock The surprising effectiveness of ppo in cooperative multi-agent games.
\newblock \emph{Advances in Neural Information Processing Systems},
  35:\penalty0 24611--24624, 2022.

\bibitem[Zheng et~al.(2018)Zheng, Yang, Cai, Zhou, Zhang, Wang, and
  Yu]{zheng2018magent}
Lianmin Zheng, Jiacheng Yang, Han Cai, Ming Zhou, Weinan Zhang, Jun Wang, and
  Yong Yu.
\newblock Magent: A many-agent reinforcement learning platform for artificial
  collective intelligence.
\newblock In \emph{Proceedings of the AAAI conference on artificial
  intelligence}, volume~32, 2018.

\end{thebibliography}

\end{document}